\documentclass[twocolumn,5p]{elsarticle}
\pdfoutput=1
\usepackage{lineno}
\modulolinenumbers[5]

\journal{Robotics and Autonomous Systems}

\usepackage[utf8]{inputenc}
\inputencoding{utf8}
\usepackage{hyperref}
\usepackage{graphics} 
\usepackage{epsfig} 
\usepackage{amsmath} 
\usepackage{amssymb}  
\usepackage{hhline}
\usepackage{tabulary}
\usepackage{multirow}
\usepackage{bm}
\usepackage{algorithm}
\usepackage[noend]{algpseudocode}
\usepackage{tikz}
\usetikzlibrary{arrows}
\usetikzlibrary{trees,snakes,decorations.pathmorphing,decorations.markings}
\usepackage{fixltx2e}

\tikzset{
  treenode/.style = {align=center, inner sep=1.5pt, text centered,
    font=\scriptsize\sffamily},
  arn_e/.style = {treenode, circle, black, dashed, draw=black, fill=white},
  arn_n/.style = {treenode, circle, black, draw=black,
    fill=cyan, minimum width=1.5em, minimum height=1.5em},
  arn_a/.style = {treenode, rectangle, black, draw=black, fill=red!20,
    text width=1.5em, minimum width=1.5em, minimum height=1.5em},
  arn_ch/.style = {treenode, rectangle, black, draw=black, fill=orange,
    text width=1.5em, minimum width=1.5em, minimum height=1.5em},
  arn_nch/.style = {treenode, circle, black, draw=black, fill=orange,
    minimum width=1.5em, minimum height=1.5em}
}

\DeclareMathOperator*{\argmax}{arg\,max}
\algdef{SE}[DOWHILE]{Do}{doWhile}{\algorithmicdo}[1]{\algorithmicwhile\ #1}

\begin{document}

\begin{frontmatter}

\title{Reset-free Trial-and-Error Learning for Robot Damage Recovery}

\author{Konstantinos Chatzilygeroudis}
\ead{konstantinos.chatzilygeroudis@inria.fr}
\author{Vassilis Vassiliades}
\ead{vassilis.vassiliades@inria.fr}
\author{Jean-Baptiste Mouret\corref{mycorrespondingauthor}}
\cortext[mycorrespondingauthor]{Corresponding author}
\ead{jean-baptiste.mouret@inria.fr}
\address{Inria, Villers-lès-Nancy, F-54600, France}
\address{CNRS, Loria, UMR 7503, Vandœuvre-lès-Nancy, F-54500, France}
\address{Université de Lorraine, Loria, UMR 7503, Vandœuvre-lès-Nancy, F-54500, France}

\begin{abstract}
The high probability of hardware failures prevents many advanced robots (e.g., legged robots) from being confidently deployed in real-world situations (e.g., post-disaster rescue). Instead of attempting to diagnose the failures, robots could adapt by trial-and-error in order to be able to complete their tasks. In this situation, damage recovery can be seen as a Reinforcement Learning (RL) problem. However, the best RL algorithms for robotics require the robot and the environment to be reset to an initial state after each episode, that is, the robot is not learning autonomously. In addition, most of the RL methods for robotics do not scale well with complex robots (e.g., walking robots) and either cannot be used at all or take too long to converge to a solution (e.g., hours of learning).
In this paper, we introduce a novel learning algorithm called ``Reset-free Trial-and-Error'' (RTE) that (1) breaks the complexity by pre-generating hundreds of possible behaviors with a dynamics simulator of the intact robot, and (2) allows complex robots to quickly recover from damage while completing their tasks and taking the environment into account. We evaluate our algorithm on a simulated wheeled robot, a simulated six-legged robot, and a real six-legged walking robot that are damaged in several ways (e.g., a missing leg, a shortened leg, faulty motor, etc.) and whose objective is to reach a sequence of targets in an arena. Our experiments show that the robots can recover most of their locomotion abilities in an environment with obstacles, and without any human intervention.
\end{abstract}

\begin{keyword}
Robot Damage Recovery, Autonomous Systems, Robotics, Trial-and-Error Learning, Reinforcement Learning
\end{keyword}

\end{frontmatter}


\section{Introduction}

During the recent DARPA Robotics Challenge (2015), many robots had to be ``rescued'' by humans because of hardware failures~\cite{atkeson_no_2015}, which is paradoxical for robots that were designed to operate in environments that are too risky for humans. While these robots could certainly have been more robust  and some falls prevented, even the best robots will encounter unforeseen situations: hardware failures will always be a possibility, especially with highly complex robots in complex environments \cite{carlson_how_2005}. For instance, C. Atkeson et al. report that the Atlas robot they used in the DARPA Robotics challenge had a ``mean time between failures of hours or, at most, days''~\cite{atkeson_no_2015,dedonato2017team}.

\begin{figure}[!t]
  \centering
  \includegraphics[width=\linewidth]{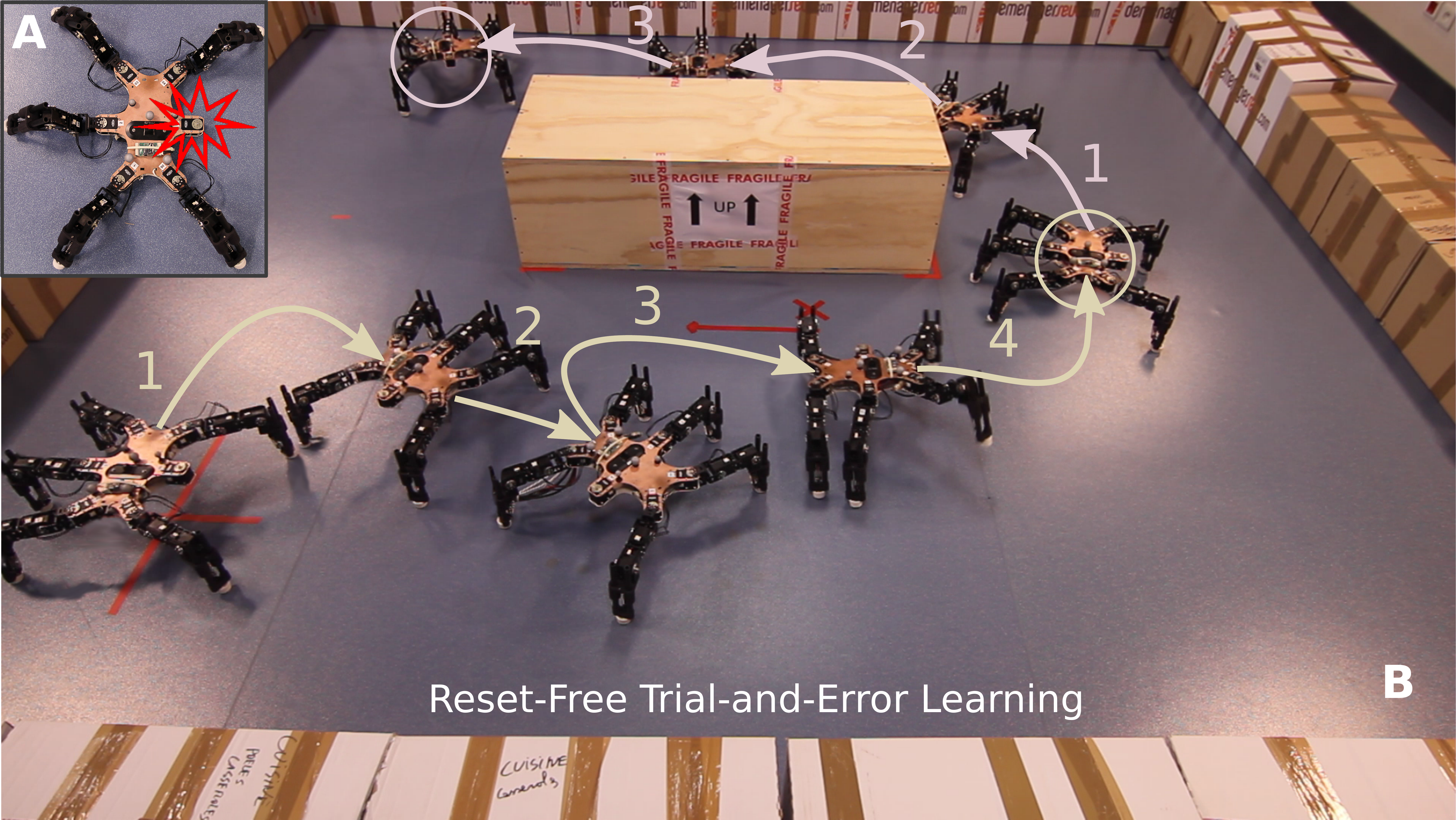}
  \caption{A typical experiment with the Reset-free Trial-and-Error (RTE) algorithm. \textbf{A.} A 6-legged (hexapod) robot is damaged; i.e., missing a leg. \textbf{B.} The robot uses RTE to learn how to compensate while completing its task and taking into account the environment. As the robot moves, it improves its performance, i.e., it needs fewer episodes to reach the next target.} 
  \label{fig:concept_ite_mcts}
\end{figure}

The traditional method for damage recovery is to first diagnose the failure, then update the plans to bypass it~\cite{isermann2006fault,verma_real-time_2004,lengagne2013generation}. Nevertheless, conceptually, the probability of failing grows exponentially with the complexity of the robot (e.g., a Roomba vs a humanoid) and the environment (e.g., an empty arena vs a post-earthquake building); accurate diagnosis, therefore, becomes much more challenging and requires many more internal sensors, which, in turn, increase the complexity and the cost of robots.

To overcome these challenges, robots can avoid the diagnosis step and directly learn a compensatory behavior by trial-and-error~\cite{cully_robots_2015,koos_fast_2013,ren2015multiple}. In this case, damage recovery is a reinforcement learning (RL) problem in which the robot has to maximize its performance for the task at hand \emph{in spite of being damaged}~\cite{kober_reinforcement_2013}.
The most successful traditional RL methods typically learn an action-value function that the agent consults to select the best action from each state (i.e., one that maximizes long-term reward)~\cite{sutton1998reinforcement,mnih_human_level_2015}. These methods work well in discrete action spaces (and even better when combined with discrete state spaces), but robots are typically controlled with continuous inputs and outputs (see~\cite{deisenroth2013survey,kober_reinforcement_2013} for detailed discussions on the issues of classic RL methods in robotics).

As a result, the most promising approaches to RL for robot control do not rely on value functions; instead, they are \emph{policy search} methods that learn parameters of a controller, called the policy, that maps sensor inputs to joint positions/torque~\cite{deisenroth2013survey}. These methods make it possible to use policies that are well-suited for robot control such as dynamic movement primitives~\cite{ijspeert2002learning} or general-purpose neural networks~\cite{levine2013guided}.
In \emph{direct} policy search, the algorithms view learning as an optimization problem that can be solved with gradient-based or black-box optimization algorithms~\cite{stulp_robot_2013}. As they are not modeling the robot itself, these algorithms scale well with the dimensionality of the state space. They still encounter difficulties, however, as the number of parameters which define a policy, and thus the dimensionality of the search space, increases~\cite{deisenroth2013survey}.
In \emph{model-based policy search}, the algorithms typically alternate between learning a model of the robot and learning a policy using the learned model~\cite{deisenroth_gaussian_2015,chatzilygeroudis2017black}. As they optimize policies without interacting with the robot, these algorithms not only scale well with the number of parameters, but can also be very \emph{data efficient}, requiring few trials on the robot itself to develop a policy. They do not scale well with the dimensionality of the state space, however, as the complexity of the dynamics tends to scale exponentially with the number of moving components.

In addition to scaling, another limitation of most of the current RL methods used in robotics is that after each trial, the robot needs to be reset to the same state~\cite{kober_reinforcement_2013,deisenroth_learning_2011}. While this reset is often not a problem for a manipulator, it prevents mobile robots (e.g., a stranded mobile manipulator or a legged robot) from exploiting this kind of algorithms to recover from damage in real-world situations. The robot cannot ignore its environment while learning, which is usually the case, as it may be further damaged if it makes a wrong decision. For example, if the robot is in front of a wall and needs to try a new way to move, it should not try to go forward, but it should select actions that would make it more likely to move backwards in order to avoid hitting the wall.

An ideal damage recovery algorithm should therefore (1) not need any reset between episodes, (2) scale well enough 
with respect to the dimensionality of the state/action space of the robot, so that it can be used for ``complex'' robots (e.g., legged robots) with the computing resources that are typically embedded in modern robots, and (3) explicitly take into account the environment. \emph{The objective of the present paper is to introduce a reinforcement learning algorithm that fits these three requirements by exploiting specific features of the damage recovery problem.}

More precisely, we investigate a simplified scenario that captures these challenges: a waypoint-controlled robot is damaged in a way that is unknown to its operator (e.g., a leg is partially cut or a motor working at half speed); to get out of the building, the robot must recover its locomotion abilities so that it can reach the waypoints fixed by its operator. Our objective is to have the robot recover its locomotive abilities to the maximum extent possible in the shortest amount of time (Fig.~\ref{fig:concept_ite_mcts}). We assume that no diagnosis is available or that the diagnosis failed, either because the robot lacks the right sensor or because the damage is so out of the ordinary that it cannot be properly diagnosed. For simplicity, we also assume that the environment is known to the robot; we will discuss possible extensions of our approach when the environment is unknown in the discussion section.

Our first source of inspiration is the recently introduced Intelligent Trial and Error (IT\&E) algorithm~\cite{cully_robots_2015}. This algorithm is an episodic policy search algorithm that is specifically designed for damage recovery. It addresses the scaling challenge by assuming that some high-performing policies for the intact robot still work on the damaged robot. While this assumption does not always hold, empirical experiments show that it often holds with highly redundant robots (e.g., legged robots or humanoids)~\cite{cully_robots_2015,koos_fast_2013} because (1) there are often many ways to perform a task, and (2) the outcomes of behaviors that do not use the damaged parts are similar between the intact and the damaged robot. Using this assumption, IT\&E searches for a diverse set of high-performing policies \emph{before the mission} (offline), then performs the \emph{online} search, that is, the adaptation to damage, by searching solely in this lower-dimensional set of pre-selected policies (using Bayesian optimization)~\cite{cully_robots_2015}.
As a result, most of the trials required for the policy search are transferred from the real damaged robot, which can perform only a few trials, to simulations with the intact robot, which can perform many more trials, especially on modern computing clusters. For instance, IT\&E allows an 18-DOF hexapod robot to learn to walk after several injuries within a dozen episodes~\cite{cully_robots_2015} and only two minutes of combined interaction and computation time.

%
A second source of inspiration is the recent AlphaGo algorithm that succeeded in beating the European and World champions in the game of Go~\cite{silver2016mastering}. Essentially, the authors use deep learning to pre-compute default policies and initial values for a Monte Carlo Tree Search (MCTS)~\cite{chaslot_monte-carlo_2008} algorithm that plans (approximately) the best next action to take. We can draw an analogy in robotics and pre-compute actions or policies, learn the model of the robot on-line (the physical robot is damaged) and use MCTS to select the most promising action.
Interestingly, MCTS can also take into account the uncertainty of the prediction of the model of the environment (e.g., when using Gaussian processes for models~\cite{nguyen2011model}). Unfortunately, it seems unrealistic to learn a probabilistic model of the full dynamics of a walking robot (like in~\cite{hester2013texplore}) within a few seconds (or minutes) of interaction time and the on-board computational power of a typical robot; more importantly, a probabilistic planner that would plan in the full controller space is even more computationally demanding.

Our main idea is to adapt the pre-computing part of IT\&E, so that it can be used by a MCTS-based planner to select the next trial, in place of the Bayesian optimization used in IT\&E. In addition, we utilize a probabilistic model to learn how to correct the outcome of each action on the damaged robot and use the MCTS-based planner in a similar way as in AlphaGo~\cite{silver2016mastering} and the TEXPLORE algorithm~\cite{hester2013texplore}, but also incorporating the uncertainty of the model prediction in the search. This allows us to propose a trial-and-error learning algorithm for damage recovery that can work on a real hexapod robot, within reasonable computation time (less than 1 minute between each episode), that does not need any reset between each episode and takes into account the environment when learning. We call this new algorithm ``Reset-free Trial-and-Error'' (RTE). In this paper, we will show that RTE performs significantly better than a modified (improved) version of TEXPLORE in both a simple differential drive mobile robot and a hexapod robot locomotion task (in the latter task, we empirically evaluate that TEXPLORE is not applicable due to the dimensionality of the action space).
\begin{figure*}[!t]
  \centering
  \vspace*{5pt}
  \includegraphics[width=\linewidth]{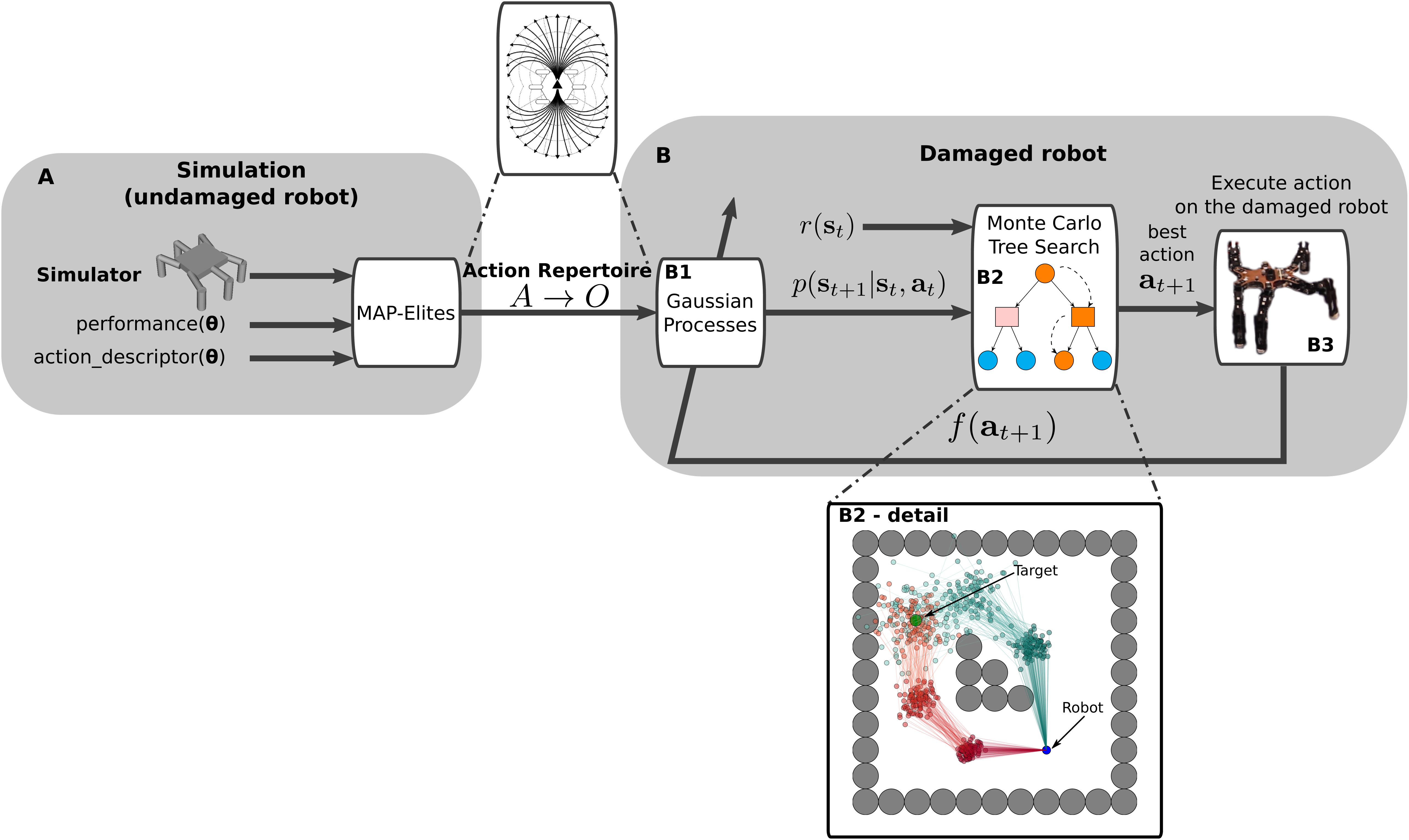}
  \caption{Overview of Reset-free Trial-and-Error (RTE) algorithm. \textbf{A.} Before deploying the robot, simulations with the intact robot are used to generate an action repertoire with the MAP-Elites algorithm. \textbf{B.} This repertoire is refined using a probabilistic learning model (Gaussian processes here --- \textbf{B1}). This model is then used as the black-box simulator of a probabilistic planner (Monte Carlo Tree Search here --- \textbf{B2}), which computes and outputs the best action to complete the task taking into account the uncertainty of the model. To better illustrate what happens in this phase, we \emph{``zoom in''} and illustrate one simple case (\textbf{B2-detail}). The robot (blue circle) has to reach the target (green circle) without hitting the obstacles (gray circles) and its model is uncertain. The algorithm explores several alternative paths to the goal (here only 2 for illustration purposes) and chooses the path that achieves the largest expected return (here we select the red one, as the green one collides more often). The lines of the same color are sampled from the distribution of choosing the specific action sequence. Once the best path is selected, the physical damaged robot executes the first action (\textbf{B3}) of the path and updates the repertoire with the new gathered data. The algorithm then re-explores new ways to reach the goal 
  and the process continues until the task is completed.}
  \label{fig:concept_detail}
\end{figure*}
%
%
%
%

The main contributions of this paper are as follows:
\begin{itemize}
  \item a novel formulation of robot damage recovery as a model-based RL problem;
  \item a novel combination of learning techniques that resembles that of AlphaGo and exploits simulations of the intact robot to accelerate learning on the physical, damaged robot;
  \item extensive experiments in simulation with a damaged simple differential drive mobile robot and a damaged hexapod (6-legged) robot, which validate the performance of the proposed approach and show that RTE performs and scales significantly better than TEXPLORE;
  \item experimental validation on a physical, damaged hexapod robot that recovers most of its locomotion abilities and is able to complete its task(s), \emph{without any human intervention}.
\end{itemize}

%
\section{State of the art}

\subsection{Learning in Robotics}
While there is a consensus that robots should be able to learn new tasks and improve their performance over time, there is far less agreement on the best place to insert learning in a robot learning architecture. Many approaches rely on supervised learning algorithms that learn forward or inverse models. Once learned, such models can be combined with control or planning algorithms to achieve the task at hand. A critical aspect of model learning is data acquisition: supervised learning algorithms need labeled data, but random babbling is often insufficient to generate behaviors that are interesting for robots~\cite{nguyen2011model,baranes2013active}; for instance, random movements with a robotic arm are unlikely to generate grasp-like behaviors. Active learning can alleviate this issue by exploring behaviors that improve the model ``at the right place''~\cite{baranes2013active}.

%
Instead of learning a model, robots can use an RL algorithm to discover how to behave~\cite{kober_reinforcement_2013}. Classic RL approaches, however, are designed for discrete state and action spaces~\cite{sutton1998reinforcement,kober_reinforcement_2013}, whereas robots almost always have to solve continuous tasks, for example balancing by controlling joint torques~\cite{nori2015icub}.
A popular alternative is to view RL as an optimization of the parameters of a policy~\cite{stulp_robot_2013}, which can be solved with gradient-based methods~\cite{peters2008reinforcement,kober_reinforcement_2013}, evolutionary algorithms~\cite{2012ACLI2061} or other optimization methods like Bayesian optimization~\cite{cully_robots_2015,calandra2015bayesian,Lizotte2007}. Nevertheless,
most policy search algorithms require a reset of the environment and robot, or specific initial states. A few non-episodic policy search algorithms (that lift these constraints) exist in the literature~\cite{montgomery2016rfgps,tedrake2004stochastic,peters2010relative,schulman2015trust}. 
However, even for simple problems like reaching targets in 2D with a 3-DOF arm, they still typically require a few hundred samples~\cite{montgomery2016rfgps}.

Several algorithms combine ideas from model learning and from RL. In particular, TEXPLORE is a non-episodic model-based RL algorithm that is based on learning the transition dynamics of the robot, which are then used by an MCTS-based planner to select the most promising action in the current time step or episode~\cite{hester2013texplore}. The anytime nature of MCTS allows TEXPLORE to stop the planning procedure when required and thus can even be used for real-time control. For example, TEXPLORE has successfully been used to control a real car in real-time~\cite{hester2012rtmba}. Nevertheless, TEXPLORE has only been used with deterministic models and with discrete action spaces. Moreover, as the dimensionality of the action space increases, the performance of MCTS rapidly deteriorates~\cite{browne2012survey}, preventing the use of TEXPLORE in more complex robots. Finally, learning a full dynamics model of a complex robot (as empirically evaluated in this paper) cannot be done in a data-efficient manner~\cite{droniou2012learning}.

\subsection{Fault Tolerance and Recovery in Robotics}
\label{sec:sota}
Classic approaches to fault tolerance rely on updating the model of the robot, either directly with self-diagnosis~\cite{blanke2003diagnosis}, or indirectly with machine learning~\cite{bongard2006resilient,verma_real-time_2004}; the model is then used for planning and/or control. For instance, if a hexapod robot detects that one of its legs is not working as expected, it can stop using it and adapt the controller to use only the working legs~\cite{mostafa2010alternative}. However, because of the many perceptual ambiguities on a robot, these approaches need many sensors and/or strong hypotheses about the kind of possible faults.

An alternative approach is to let a damaged robot learn a new policy by trial-and-error, which bypasses the need for a diagnosis~\cite{koos_fast_2013,cully_robots_2015,ren2015multiple}. In this line of work, the biggest challenge is to design algorithms that are as data-efficient as possible, because too many trials may damage the robot further, and learning must be rapid enough to be useful in real-world situations. To minimize the number of trials, several algorithms rely on the transferability hypothesis~\cite{cully_robots_2015,koos_fast_2013}: the behaviors that do not use the damaged parts are likely to be similar between the damaged and the undamaged robot, therefore \emph{simulations of the intact robot can help when searching for a new behavior on the damaged robot}. Starting with this hypothesis, the IT\&E algorithm~\cite{cully_robots_2015} exploits a dynamic simulation of the \emph{intact} robot to create a behavior-performance map that predicts the performance of thousands of different behaviors. If damage occurs, this map is used as a prior for a Bayesian optimization algorithm that searches for a compensatory behavior~\cite{shahriari2016taking,Lizotte2007,calandra2015bayesian}. Overall, the experimental results show that IT\&E can allow various types of robots (a hexapod robot and an 8-DOF arm) to compensate for many different injuries in less than 2 minutes~\cite{cully_robots_2015}.
%
%
\subsection{Probabilistic and Sample-Based Planning}

Sample-based planning is one of the main philosophies that addresses the motion planning problem~\cite{lavalle2006planning}. The traditional algorithms in this category are ``Rapidly Exploring Random Trees'' (RRT)~\cite{lavalle1998rapidly} and ``Probabilistic Roadmaps for path planning in high-dimensional configuration spaces'' (PRM)~\cite{kavraki1996probabilistic}. In RRT, a tree is constructed incrementally from samples drawn randomly from the search space and is biased to grow towards big unexplored areas of the problem. The basic idea behind PRM is to take random samples from the configuration space of the robot, test if they are in the free space, and then use a local planner to attempt to connect these configurations to other nearby configurations.

A more recent algorithm that belongs to this category is Monte Carlo Tree Search (MCTS)~\cite{chaslot_monte-carlo_2008}. MCTS is a method for finding optimal decisions in a given domain by taking random samples in the decision space and building a search tree according to the results. It has already had a profound impact on Artificial Intelligence approaches for domains that can be represented as trees of sequential decisions, particularly games and planning problems~\cite{browne2012survey,silver2016mastering}.

\section{Problem Formulation}
\label{sec:problem_form}
Our problem can be cast in the general framework of Markov Decision Processes (MDP)~\cite{sutton1998reinforcement}. An MDP is a tuple $(S,A,T,r)$, where $S$ is the state space (continuous or discrete), $A$ is the action space (continuous or discrete), $T(\mathbf{s}_{t},\mathbf{a}_{t},\mathbf{s}_{t+1})$ is the state transition function specifying the probability of transitioning to state $\mathbf{s}_{t+1} \in S$
when the agent takes action $\mathbf{a}_{t} \in A$ in state $\mathbf{s}_{t} \in S$, and $r: S \rightarrow \mathbb{R}$ is the immediate reward function (which defines the task of the agent), with $r(\mathbf{s}_{t+1})$ being the immediate reward of state $\mathbf{s}_{t+1}$ and $\mathbf{s}_{t+1}$ may contain both internal variables (such as body position) and external variables (such as obstacles).
The objective of the agent (i.e., the robot) is to find a deterministic policy $\pi$, i.e., a mapping from states to actions, $\mathbf{a}_t = \pi(\mathbf{s}_t)$, that maximizes its expected discounted return:
\begin{equation}
J^{\pi} = \mathbb{E} \Bigg[ \sum_{t=0}^{\infty} \gamma^{t}r(\mathbf{s}_{t+1}) \Big| \pi \Bigg]
\end{equation}
where $\gamma \in [0,1)$ is a factor that discounts future rewards.
%
$T$ and $r$ describe the environmental dynamics and they are collectively known as the model of the environment.
If the agent has access to this model, it can use a planning algorithm to find the optimal policy. In this paper, the transition function $T$ is learned and we assume that the reward function $r$ is known to the robot.

In our setting, the robot needs to execute a sequence of related tasks $G_1, G_2, \dots, G_n$, each of which is a shortest path problem:
%
\begin{equation}
r(\mathbf{s}_{t}) =
\left\{
\begin{array}{ll}
R_{goal}  & \mbox{if } \mathbf{s}_{t} = goal(G_i) \\
-R_{term} & \mbox{if } \mathbf{s}_{t} = terminal(G_i) \\
0 & \mbox{otherwise}
\end{array}
\right.
\end{equation}
where $R_{goal}>0, R_{term}\geq 0$, $goal(G_i)$ returns the goal state of task $G_i$, and $terminal(G_i)$ returns a non-goal, terminal state of task $G_i$, e.g., a colliding state. When $\mathbf{s}_{t} = goal(G_i)$, the robot finishes task $G_i$ and starts executing task $G_{i+1}$.

\section{Approach}
\subsection{Overview}
\label{sec:approach}
RTE allows robots to ``learn while doing'' instead of ``learning and then doing''. This is achieved by:
\begin{itemize}
  \item pre-computing an action repertoire with relatively low-fidelity simulations (e.g., perfect velocity actuators) of the intact robot (generated by MAP-Elites~\cite{mouret_illuminating_2015}, Fig.~\ref{fig:concept_detail}A) that also \textbf{(a)} creates a mapping between the task space and the parameters of the low-level controller and \textbf{(b)} reduces the dimensionality of the action space;
  \item using a probabilistic model (Gaussian processes) to learn how to correct the prediction of the outcome of each action for the damaged robot (Fig.~\ref{fig:concept_detail}B1);
  \item re-planning at every episode with a probabilistic planner (Monte Carlo Tree Search) that selects the next action to execute, based on the predictions of the probabilistic model, the uncertainty of those predictions, the environment, the current state of the robot, and the target state (Fig.~\ref{fig:concept_detail}B2). More specifically, we solve a path/motion planning problem with uncertain transitions (Fig.~\ref{fig:concept_detail}B2-detail). Clearly, the further we plan into the future, the more uncertain our estimates will be about where the robot will end up (Fig.~\ref{fig:concept_detail}B2-detail); therefore, an ideal planner would select the action that has the best utility (in terms of expected discounted cumulative reward) by considering these future estimates (i.e., how close they arrive to the target, how often they hit obstacles).
\end{itemize}

In summary, if damage occurs, RTE performs the following loop (Fig.~\ref{fig:concept_detail}B): (1) uses MCTS to select the next best action from the repertoire to complete the task, (2) executes the action for a given time duration (e.g., 3 seconds or 100 simulation steps), that is, \emph{perform an episode}, (3) updates the Gaussian processes (GPs) to improve the prediction of the outcome of each action of the repertoire and (4) repeats (1)-(3) until the task(s) are completed.

\subsection{Learning the Action Repertoire}
\label{sec:map-elites}
Controllers for complex robots, for instance legged robots, usually involve numerous parameters, which makes control policies challenging to learn within a few trials. We circumvent this issue by using the transferability hypothesis (Sec.~\ref{sec:sota}) and learn, before deploying the robot, a repertoire of controllers with a simulated intact robot. The predicted outcomes of the actions will be refined online after each action is executed (i.e., at the end of each episode) by the damaged robot (Sec.~\ref{sec:gp}).
\begin{algorithm}
  \small
  \caption{MAP-Elites}\label{algo:map_elites}
  \begin{algorithmic}[1]
    \Procedure{MAP-Elites}{}
      \State ($\mathbf{P} \gets \emptyset, \mathbf{\Theta} \gets \emptyset$)
      \Comment{\emph{Performance and feature grids}}
      \For {$i=1\to G$} \Comment{\emph{Initialization: $G$ random $\theta$}}
        \State $\boldsymbol{\theta} = $ random\_solution()
        \State add\_to\_repertoire($\boldsymbol{\theta}, \mathbf{P}, \mathbf{\Theta}$)
      \EndFor
        \For {$i=1\to I$} \Comment{\emph{Main loop, $I$ iterations}}
        \State $\boldsymbol{\theta} = $ random\_selection($\mathbf{\Theta}$)
        \State $\boldsymbol{\theta}' = $ random\_variation($\boldsymbol{\theta}$)
        \State add-to-repertoire($\boldsymbol{\theta}', \mathbf{P}, \mathbf{\Theta}$)
      \EndFor
      \Return repertoire and performance ($\mathbf{\Theta}$, $\mathbf{P}$)
    \EndProcedure

    \Procedure{add-to-repertoire}{$\boldsymbol{\theta}, \mathbf{P}, \mathbf{\Theta}$}
      \State $\mathbf{a} = $ action\_descriptor($\boldsymbol{\theta}$) \Comment{\emph{Use the forward model}}
      \State $p = $ performance($\boldsymbol{\theta}$) \Comment{\emph{Use the forward model}}
      \If {$\mathbf{P}(\mathbf{a}) = \emptyset$ or $\mathbf{P}(\mathbf{a}) < p$} \Comment{\emph{Replace if better}}
      \State $\mathbf{P}(\mathbf{a}) = p$
      \State $\mathbf{\Theta}(\mathbf{a}) = \boldsymbol{\theta}$
      \EndIf
    \EndProcedure
  \end{algorithmic}
\end{algorithm}

We assume that the robot is controlled by a low-level controller that is parametrized by a vector $\boldsymbol{\theta} \in \mathbb{R}^{d}$. 
We also assume that each point in the task space can be described by a vector $\mathbf{a} \in \mathbb{R}^{n_a}$, which we call an ``action descriptor''.
We would like to create a repertoire that covers the task space as well as possible~\cite{cully_robots_2015,cully_evolving_2015,duarte2017evolution}, i.e., to both determine a good set of actions $A$ and a mapping between $A$ and $\Theta$ ($A \rightarrow \Theta$).
This mapping also reduces the dimensionality of the search space since the task space is usually much lower dimensional than the controller space.

If we take a robotic manipulator as an example, the controller space could be joint positions, the task space could be the $(x,y,z)$ coordinates of the end-effector, and the repertoire will map $(x,y,z)$ positions to joint positions, that is, it would be a discrete representation of the inverse kinematics of the arm. Nonetheless, while an inverse kinematics solver could be used to create a repertoire for a manipulator, most robots do not have access to such inverse models. This is true for walking robots, in particular.

As a consequence, instead of using an inverse model, we learn the action repertoire with an iterative algorithm called MAP-Elites~\cite{mouret_illuminating_2015,cully_robots_2015} and a forward model (e.g., a dynamic simulator). As with the inverse kinematics of redundant manipulators, the mapping from the parameter space to the task space is typically many-to-one. Thus, we need to define a performance function to select the best $\boldsymbol{\theta}$ for each point of the task space.
This performance function is designed so as to promote certain type of behaviors (Sec.~\ref{sec:controller}) and does not coincide with the reward function of the MDP.

Essentially, MAP-Elites discretizes the $n_a$-dimensional task space to an $n_a$-dimensional grid, and then attempts to fill each of the cells using a variation-selection loop~\cite{mouret_illuminating_2015,cully_robots_2015,cully2017quality}. Algorithmically, it starts with $G$ random parameter vectors, simulates the robot with these parameters, and records both the position of the robot in the task space and the performance (Algo. \ref{algo:map_elites}, 3-5). If the cell is free, then the algorithm stores the parameter vector in that cell; if it is already occupied, then the algorithm compares the performance values and keeps only the best parameter vector (Algo. \ref{algo:map_elites}, 10-15). Once this initialization is done, MAP-Elites iterates a simple loop (Algo. \ref{algo:map_elites}, 6-9): (1) randomly selects one of the occupied cells, (2) adds a random variation to the parameter vector, (3) simulates the behavior, (4) inserts the new parameter vector into the grid if it performs better or end-ups in an empty cell (discard the new parameter vector otherwise).

While MAP-Elites is computationally expensive, it can be straightforward to parallelize and can run on large clusters before deploying the robot. So far, it has been successfully used to generate: behaviors for legged robots~\cite{cully_robots_2015}, robotic arms~\cite{cully_robots_2015, mouret_illuminating_2015} and wheeled robots~\cite{duarte2016evorbc,pugh2016quality,duarte2017evolution}; designs for airfoils~\cite{gaier2017feature}, as well as for the morphologies of walking ``soft robots''~\cite{mouret_illuminating_2015}; adversarial images for deep neural networks~\cite{nguyen2015deep}; ``innovation engines'' which generate images that resemble natural objects~\cite{nguyen2016understanding}; and 3D-printable objects using feedback from neural networks trained on 2D images~\cite{lehman2016iccc}.
MAP-Elites has also been extended to effectively handle task spaces of arbitrary dimensionality~\cite{vassiliades2017using}.

\subsection{Learning with Gaussian Processes}
\label{sec:gp}
MAP-Elites provides not only the set of actions to be used by the planner, but also a prior on how an action modifies the state variables, i.e., a mapping from actions to relative outcomes, $f: A\to O$.
%
%
Since this prior comes from a simulator and the simulator uses a model of the intact robot, it is only an approximation.
Therefore, to make the physical damaged robot perform well, there needs to be a way to correct this mapping.

To do so, we use $n$ Gaussian Processes (where $n$ is the number of dimensions of $O$) with a mean function that corresponds to the prior provided by MAP-Elites. In other words, the mapping computed with the simulator serves as a prior for the GPs.

A GP is an extension of the multivariate Gaussian distribution to an infinite-dimension stochastic process for which any finite combination of dimensions will be a Gaussian distribution~\cite{rasmussen2006gaussian}. For each dimension $d = 1\dots n$, it is a distribution over functions, specified by its mean function, $\mu_d(\cdot)$ and covariance function, $k_d(\cdot,\cdot)$:
\begin{equation}\label{eq:f_gp_full}
f_d(\mathbf{a}) \sim GP(\mu_d(\mathbf{a}), k_d(\mathbf{a}, \mathbf{a}'))
\end{equation}
Assuming $D^d_{1:t} = \{f_d(\mathbf{a_1}),\cdots,f_d(\mathbf{a_t})\}$ is a set of observations, $M_d(\cdot)$ is the mean from the simulated prior and $\sigma_w^2$ the sampling noise, the GP is computed as follows:
\begin{align}\label{eq:gp_mean}
p(f_d(\mathbf{a})|D^d_{1:t}, \mathbf{a}) = \mathcal{N}(\mu_d(\mathbf{a}), \sigma_d^2(\mathbf{a}))\\
\mu_d(\mathbf{a}) = M_d(\mathbf{a}) + \boldsymbol{k^\top}_d(\boldsymbol{K}_d+\sigma_w^2I)^{-1}(D^d_{1:t} - M_d(\mathbf{a}_{1:t}))\\
\sigma_d^2(\mathbf{a}) = k_d(\mathbf{a},\mathbf{a}) - \boldsymbol{k^\top}_d(\boldsymbol{K}_d+\sigma_w^2I)^{-1}\boldsymbol{k}_d
\end{align}
%
where $\boldsymbol{K}_d$ is the kernel matrix with entries $K^{ij}_d = k_d(\mathbf{a}_i,\mathbf{a}_j)$ and $\boldsymbol{k}_d = k_d(D^d_{1:t},\mathbf{a})$.

\subsection{Probabilistic Optimal Planning using MCTS}
\label{sec:mcts}

At the end of each episode, we need to solve an MDP with an action set that contains thousands of actions in a continuous state space. Since GPs are probabilistic models, they provide both a prediction and the uncertainty associated with each prediction, which can be exploited by probabilistic planners. Here we use Monte Carlo Tree Search (MCTS)~\cite{chaslot_monte-carlo_2008}, as it has already been successfully used to solve (Partially Observable)-MDPs with stochastic transition functions~\cite{silver2010monte,browne2012survey}, continuous state spaces, and high branching factors~\cite{browne2012survey,couetoux2011continuous}.
%
%
%

MCTS is a best-first, sample-based search algorithm for finding optimal decisions in a given domain by taking random samples in the decision space and building a search tree according to the results. Every state in the search tree is evaluated by the average outcome of Monte Carlo rollouts from that state. These rollouts are typically random or directed by a simple, domain-dependent heuristic~\cite{browne2012survey}.
%
%
\begin{figure}[!t]
  \centering
  \vspace*{5pt}
  \includegraphics[width=0.8\linewidth]{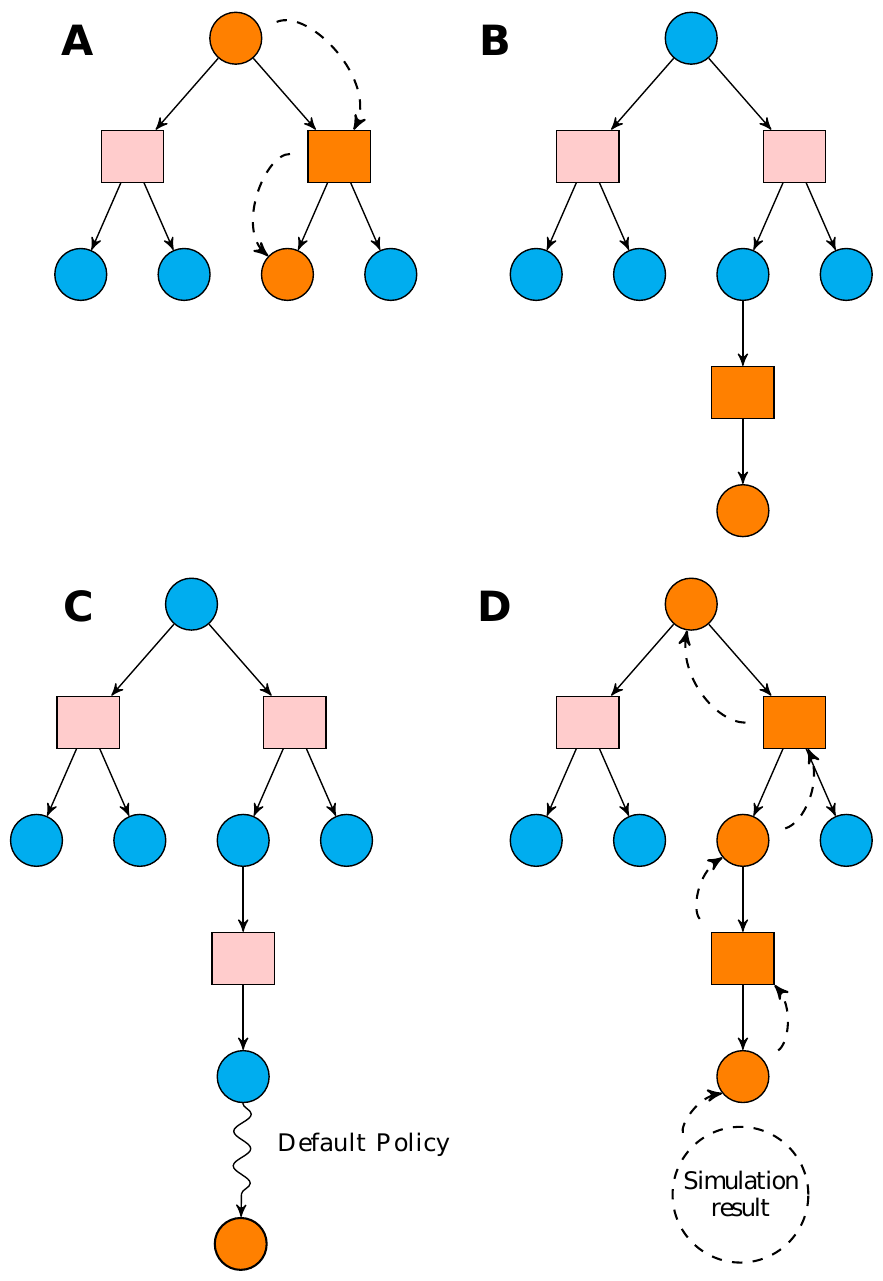}
  \caption{The four generic steps of Monte Carlo Tree Search algorithms. The circular nodes represent \emph{decision} nodes (states from where actions are selected) and the rectangular nodes represent \emph{random} nodes (state-action pairs where random outcomes can happen). See~\cite{couetoux2011continuous} for further details. \textbf{A.} The most urgent expandable node (i.e., one with no previous visits) is selected using a selection policy. \textbf{B.} The tree is expanded according to the available actions. \textbf{C.} A rollout is performed from the new node according to the default policy. \textbf{D.} The rollout result is ``backed up'' through the selected nodes.}
  \label{fig:mcts}
\end{figure}

\begin{algorithm}
\small
\caption{Generic Monte Carlo Tree Search}\label{algo:mcts}
\begin{algorithmic}[1]
\Procedure{MCTS-Search}{$\mathbf{s}_0$}
  \While {within computational budget}
  \State $\mathbf{s} = \mathbf{s}_0$
  \Do
  \State $\mathbf{a} =$ SelectionPolicy($\mathbf{s}$)
  \State $Children(\mathbf{s}) = Children(\mathbf{s})\cup(\mathbf{s},\mathbf{a})$
  \State $(\mathbf{s}',\rho) =$ ExpansionPolicy($\mathbf{s}, \mathbf{a}$) \Comment{\emph{see~\cite{couetoux2011continuous}}}
  \State $Children(\mathbf{s}, \mathbf{a}) = Children(\mathbf{s}, \mathbf{a})\cup \mathbf{s}'$
  \State $R(\mathbf{s},\mathbf{a}) = \rho$
  \State $\mathbf{s} = \mathbf{s}'$
  \doWhile {$n(\mathbf{s})>0$ and $\mathbf{s}$ not a terminal state} \Comment{\emph{$n(\cdot)$ returns the number of visits of a state}}
  \State $\Delta =$ Rollout($\mathbf{s}$)  \Comment{\emph{Use GPs (Sec.~\ref{sec:gp},~\ref{sec:mcts_implement})}}
  \State BackUp($\mathbf{s},\Delta, R$)
  \EndWhile
  \Return BestChild($\mathbf{s}_0$)
\EndProcedure
\end{algorithmic}
\end{algorithm}

MCTS (Algo.~\ref{algo:mcts}) is an anytime planning algorithm, i.e., it runs until some predefined computational budget (typically, a time, memory or iteration constraint) is reached, at which point the search is halted and the best-performing root action is returned. Four steps are applied per search iteration:

\begin{itemize}
  \item \emph{SelectionPolicy:} Starting at the root node, a child selection policy is recursively applied to descend through the tree until the most urgent expandable node is reached (Fig.~\ref{fig:mcts}A).
  \item \emph{ExpansionPolicy:} One child node, along with the state's associated reward $\rho = r(\boldsymbol{s})$, is added to expand the tree, according to the available actions (Fig.~\ref{fig:mcts}B).
  \item \emph{Rollout:} A rollout is performed from the new node according to the default policy to get an estimate value for this node, $\Delta$ (Fig.~\ref{fig:mcts}C). We do this by constructing a generative model using the prediction of the GPs.
  \item \emph{BackUp:} The rollout result is ``backed up'' through the selected nodes to update their statistics (Fig.~\ref{fig:mcts}D).
\end{itemize}

\subsection{Reset-free Trial-and-Error Learning Algorithm}
Connecting all the pieces together, RTE first generates an action repertoire with the MAP-Elites algorithm (Algo.~\ref{algo:mcts_ite}, lines 2-3); then, while in mission, it re-plans at each episode using MCTS and the current belief of the outcome of the actions (prediction of the GPs), taking into account the uncertainty of the predictions and potential final states (e.g., collisions with obstacle) (lines 9-13); at the end of each episode, the GPs are updated with the recorded data (lines 14-15).
\begin{algorithm}
\small
\caption{Reset-free Trial-and-Error Learning}\label{algo:mcts_ite}
\begin{algorithmic}[1]
\Procedure{RTE}{}
  \State {\small Create Action Repertoire, $A$, with MAP-Elites (Sec.~\ref{sec:map-elites})}
  \State {\small Construct mean function $M$ from MAP-Elites data}
  \For{$i=1\to$ dim$(O)$}
  \State $GP_{i}: A\to O_i$ with $M_i$ as prior (Sec.~\ref{sec:gp},~\ref{sec:gp_impl})
  \EndFor
  \While {{\bf in mission} and stopping criteria not met}
  \State RTE-EPISODE($t$)
  \State $t=t+1$
  \EndWhile
\EndProcedure
\Procedure{RTE-Episode}{$t$}
  \State $\mathbf{s}_t = $ state of robot at time $t$
  \State $\mathbf{a}_{t+1} =$ MCTS-SEARCH$(\mathbf{s}_t)$ (Sec.~\ref{sec:mcts},~\ref{sec:mcts_implement})
  \State $f(\mathbf{a}_{t+1}) =$ execute\_action$(\mathbf{a}_{t+1})$ \Comment{\emph{Execute the action and observe its outcome}}
  \State $D_{1:t+1} = \{D_{1:t}, f(\mathbf{a}_{t+1})\}$
  \For{$i=1\to$ dim$(O)$}
  \State Update $GP_i$ using $D^i_{1:t+1}$ (Sec.~\ref{sec:gp})
  \EndFor
\EndProcedure
\end{algorithmic}
\end{algorithm}
%
\section{Experimental Setup}

We investigate the following scenario: a waypoint-controlled robot is damaged in a way that is unknown to its operator (e.g., a leg is partially cut or a motor working at half speed); to get out of the building, the robot must recover its locomotion abilities so that it can reach the waypoints fixed by its operator. As already stated, we assume that no diagnosis is available or that the diagnosis failed. In addition, for the sake of simplicity, the environment is known to the robot and the robot knows its position (via a Motion Capture system). The robot has to reach 30 equidistant target waypoints in an arena with obstacles. We perform these experiments with a differential drive robot (in simulation) and with a 6-legged (hexapod) robot (in simulation and with a physical robot).

We compare three algorithms: (1) RTE, (2) a variant of RTE where the learning part is removed (i.e., MCTS-based planning with the original action repertoire --- we call this variant MCTS) and (3) a variant of TEXPLORE (we call it GP-TEXPLORE --- Algo.~\ref{algo:gp_texplore}) where: (i) the reward function is known, (ii) we use a variant of MCTS for continuous action spaces, and (iii) instead of learning the full transition dynamics, only the relative outcome of each action is learned. The main difference of GP-TEXPLORE and RTE is that the latter uses the discrete action space as defined by the learned repertoire for model learning and planning, whereas GP-TEXPLORE plans and learns the model in the full controller space. We also use GPs, without taking into account the uncertainty, instead of random forests that are used in the original TEXPLORE paper~\cite{hester2013texplore}. With (2), we try to get closer to a classic planning algorithm with re-planning after each episode.
With (3) we try to make TEXPLORE better fit our problem and we expect the original TEXPLORE algorithm to not work as well as the baseline used here. However, exploring more in these directions is beyond the scope of this paper.

\begin{algorithm}
\small
\caption{Modified TEXPLORE}\label{algo:gp_texplore}
\begin{algorithmic}[1]
\Procedure{GP-TEXPLORE}{}
  \For{$i=1\to$ dim$(O)$}
  \State $GP_{i}: \Theta\to O_i$ \Comment{controller space to outcome space}
  \EndFor
  \While {{\bf in mission} and stopping criteria not met}
  \State GP-TEXPLORE-EPISODE($t$)
  \State $t=t+1$
  \EndWhile
\EndProcedure
\Procedure{GP-TEXPLORE-Episode}{$t$}
  \State $\mathbf{s}_t = $ state of robot at time $t$
  \State $\boldsymbol{\theta}_{t+1} =$ MCTS-SEARCH$(\mathbf{s}_t)$ (Sec.~\ref{sec:mcts}) \Comment{MCTS in controller space}
  \State $f(\boldsymbol{\theta}_{t+1}) =$ execute\_action$(\boldsymbol{\theta}_{t+1})$ \Comment{\emph{Execute the action and observe its outcome}}
  \State $D_{1:t+1} = \{D_{1:t}, f(\boldsymbol{\theta}_{t+1})\}$
  \For{$i=1\to$ dim$(O)$}
  \State Update $GP_i$ using $D^i_{1:t+1}$ (Sec.~\ref{sec:gp})
  \EndFor
\EndProcedure
\end{algorithmic}
\end{algorithm}
\section{Mobile Robot Results}
\begin{figure}[!t]
  \centering
  \vspace*{5pt}
  \includegraphics[width=\linewidth]{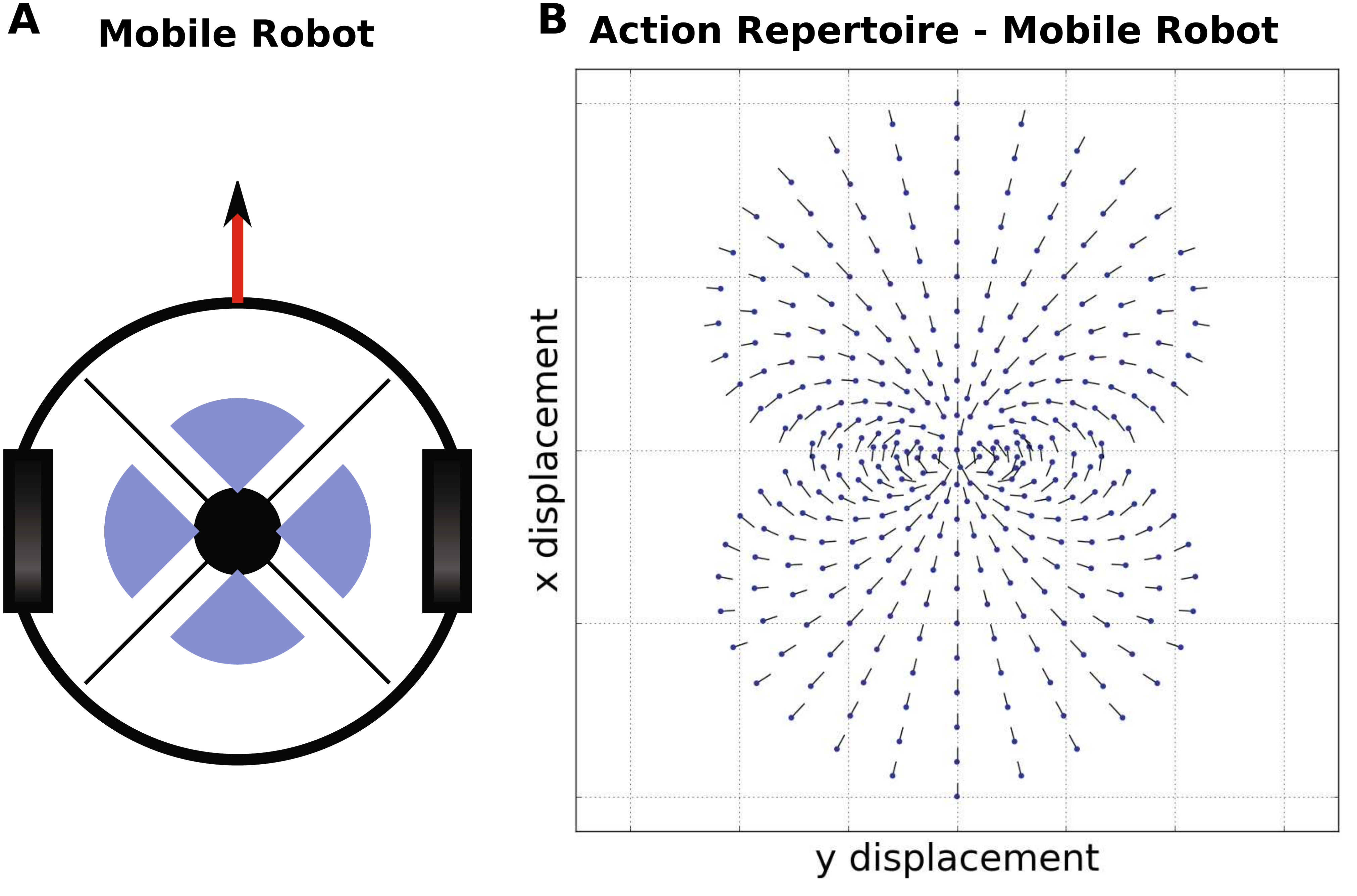}
  \caption{\textbf{A.} The velocity-actuated differential drive mobile robot used in our experiments. \textbf{B.} Repertoire for the simple robot locomotion task produced by the MAP-Elites algorithm. This repertoire maps the 2-D action descriptor (of the 3-D task space) to the 2-D controller space. Each dot represents a different action (and its $x,y$ position), while the lines indicate the orientation of the robot at the end of each behavior. All the behaviors are relative to the zero position that is located in the middle of the figure and relative to the forward orientation (line pointing up).}
  \label{fig:mobile_robot}
\end{figure}
%
The robot is a classic velocity-actuated differential drive mobile robot (Fig.~\ref{fig:mobile_robot}A). The state of the robot consists of the $(x,y)$ position and the orientation $\theta$ of the robot, i.e., $\mathbf{s}_{mob} = [x,y,\theta]$. The robot moves by applying velocities to the two wheels ($v_{left}$ and $v_{right}$).
We use the \emph{libfastsim} library for simulating the robot~\cite{2012ACLI2061}\footnote{\url{https://github.com/jbmouret/libfastsim}}. At each \emph{episode} of the learning algorithm, the velocity pair is executed for 100 time-steps (for all algorithms).
\subsection{Learning the Action Repertoire}
The robot's task is to reach points in Cartesian space $(x,y)$, therefore MAP-Elites should produce a repertoire of actions, each of which reaches a different point in the Cartesian space.
%
%
Since many controllers can reach the same position, we select those that make the robot follow a continuous-curvature trajectory and for which the body points towards the tangent of the overall trajectory at the end of the behavior. To capture this idea, we set the MAP-Elites performance of the $i_{th}$ individual to:
\begin{equation}
  p_i = \lvert\theta_i-\theta_d\rvert
\end{equation}
where $\theta_i$ is the orientation of the robot and $\theta_d$ is the desired orientation of the robot at the end of the movement.
To describe the circular trajectories we only need to keep the $(x,y)$ position of the robot at the end of the movement (since we can compute the desired angle for any point in the 2-D space). In this way we can use a 2-D action descriptor to describe the 3-D task space. The 2-D descriptor of the $i_{th}$ individual is:
\begin{equation}\label{eq:bd_desc}
\mathbf{a}_i = [
\frac{x-x_{min}}{x_{max}-x_{min}},
\frac{y-y_{min}}{y_{max}-y_{min}} ]
\end{equation}
where $x_{min}$, $x_{max}$, $y_{min}$ and $y_{max}$ are the boundaries of the reachable space ($[-100,100]$ units here).
We ran MAP-Elites for $100000$ evaluations and we got a repertoire with $331$ different actions\footnote{MAP-Elites always produces the same repertoire because the problem is easy. Note that the repertoire for such a simple robot could be generated with many other methods: here we use MAP-Elites so that we can demonstrate identical approaches for both the wheeled and the legged robot.} (Fig.~\ref{fig:mobile_robot}B). Our implementation relies on the {\it Sferes$_{v2}$}~\cite{mouret2010sferes} library.
\subsection{Learning with Gaussian Processes}
\label{sec:gp_impl}
The GP inputs are the 2-D descriptors of the actions, and the outputs are predictions of the relative $x$, $y$ and $\theta$ displacements.
To avoid angle discontinuities, instead of learning the raw angle $\theta$ we learn the $cos\theta$ and the $sin\theta$. Thus, we learn a mapping from actions to relative outcomes:
\begin{equation}\label{eq:gp_behavior} \mathbf{a} \rightarrow (\Delta x,\Delta y,cos\Delta\theta, sin\Delta\theta)\end{equation}
%
We use the {\it Squared Exponential Kernel} (SE) as the covariance function~\cite{rasmussen2006gaussian}:
\begin{equation}
  k(\mathbf{a}, \mathbf{a}')  = \sigma_{se}^2\exp \Big(-\frac{||\mathbf{a} - \mathbf{a}'||^2}{l^2}\Big)
\end{equation}
where we set $\sigma_{se}^2=0.5$ and $l=1$ in the mobile robot experiments. We also use the {\it limbo} C++11 library~\cite{cully2016limbo} for the GP regression.
\begin{algorithm}
\caption{Simple Progressive Widening}\label{algo:spw}
\begin{algorithmic}[1]
\Procedure{SPW-SelectionPolicy($\mathbf{s}$)}{}
  \If {$n(\mathbf{s})^\alpha>\#Children(\mathbf{s})$} \Comment{$0<\alpha<1$}
  \State $\mathbf{a} =$ sample\_action($\mathbf{s}$) \Comment{Sample new action from $\mathbf{s}$ (Sec.~\ref{sec:a_star})}
  \Else \Comment{Choose the action with the best UCT value~\cite{browne2012survey}}
  \State $\mathbf{a} = \argmax_{\mathbf{a}\in Children(\mathbf{s})}\hat{Q}(\mathbf{s},\mathbf{a})$, with $\hat{Q}(\mathbf{s},\mathbf{a}) = \frac{\mathbf{R}(\mathbf{s},\mathbf{a})}{n(\mathbf{s},\mathbf{a})}+ c\sqrt{\frac{ln(n(\mathbf{s}))}{n(\mathbf{s},\mathbf{a})}}$
  \EndIf
  \Return $\mathbf{a}$
\EndProcedure
\end{algorithmic}
\end{algorithm}

\subsection{Solving the problem with MCTS}
\label{sec:mcts_implement}
At the end of each episode, we need to solve an MDP with an action set that contains thousands of actions in a continuous state space and uncertain transitions (i.e., when an action is taken from the same state, the result is not the same).
In order to solve this problem, we instantiate MCTS with the following choices:
\begin{description}
\item[Selection Policy] Simple Progressive Widening (SPW --- Algo.~\ref{algo:spw})~\cite{rolet2009optimal} that properly handles cases where the action space is continuous. We set $\alpha = 0.5$  and $c = 150$.
\item[Expansion Policy] Double Progressive Widening (DPW --- Algo.~\ref{algo:dpw})~\cite{couetoux2011continuous} that properly handles cases where the state space is continuous. We set $\beta = 0.6$.
\item[Action Sampling Policy] We use A* on a simplified problem to guide the sampling procedure (Sec.~\ref{sec:a_star}).
\item[Generative Model] We construct a generative model using the prediction of the GPs:
\begin{equation}
  \label{eq:gp_gen}
  p(\mathbf{s}_{t+1} | \mathbf{s}_t, \mathbf{a}_t)\sim\mathcal{N}(\mathbf{s}_t+f(\mathbf{a}_t), \Sigma_{\mathbf{a}_t})
\end{equation}
\item[Default Policy for evaluation] Uniformly-distributed random actions from the repertoire~\cite{browne2012survey}.
\item[Best child criterion] Greedy selection, i.e., we select the action that has the maximum average cumulative reward~\cite{browne2012survey}.
\item[Reward function] $R_{goal} = 100$, reward for reaching the goal, and $R_{term} = 1000$, penalty for colliding, for each target point. We also set the reward discount factor, $\gamma=0.9$.
  \begin{itemize}
    \item For the sake of simplicity, we only used circular obstacles and a circular collision shape for the robot. Nevertheless, any shapes with the appropriate collision query functions would be compatible with our approach, since the reward function is a black-box to MCTS.
  \end{itemize}
\end{description}

\begin{algorithm}
\caption{Double Progressive Widening}\label{algo:dpw}
\begin{algorithmic}[1]
\Procedure{DPW-ExpandPolicy($\mathbf{s},\mathbf{a}$)}{}
  \If {$n(\mathbf{s},\mathbf{a})^\beta>\#Children(\mathbf{s},\mathbf{a})$} \Comment{$0<\beta<1$}
  \State Draw $s'$ from $p(\mathbf{s}'|\mathbf{s},\mathbf{a})$ \Comment{see Eq.~\ref{eq:gp_gen}}
  \State $\rho=r(\mathbf{s}')$
  \Else
  \State {\small Choose $\mathbf{s}'\in Children(\mathbf{s},\mathbf{a})$ with prob $\frac{n(\mathbf{s},\mathbf{a},\mathbf{s}')}{\sum_{\mathbf{s}_i}n(\mathbf{s},\mathbf{a},\mathbf{s}_i)}$}
  \State $\rho=r(\mathbf{s}')$
  \EndIf
  \Return $[\mathbf{s}',\rho]$
\EndProcedure
\end{algorithmic}
\end{algorithm}

To make the search faster, we implemented root parallelization in MCTS~\cite{cazenave2007parallelization} with 4 parallel trees giving a budget of $5000$ iterations to each. This implementation is available in our C++14 lightweight MCTS library\footnote{\url{https://github.com/resibots/mcts}}.

\subsection{Guiding MCTS using A* on a simplified problem}
\label{sec:a_star}
MCTS traditionally samples actions randomly. To speed up the process, we first discretize the space and create a grid map; then, we simulate a virtual point robot with 8 actions (one for each neighboring cell --- allowing diagonal moves) and solve the path planning problem using A*. Solving this simplified task requires very little computation. We use the optimized path to calculate an approximate desired direction for the next MCTS action. Next, we sample $N$ ($100$ in our case) random actions from the repertoire and return the one that best matches this direction. Note that we are using the prediction of the GPs to decide which action we should choose.
This simple procedure has the desirable effect of reducing the running time of MCTS (less than $40-50\,s$ to choose the next action), without sacrificing the quality of the returned actions.
We use this ``trick'' because our problem is path-planning, but similar tricks can be used in other problems. More generic approaches would be the {\it Blind Value} action sampling or the continuous Rapid Action Value Estimation (cRAVE)~\cite{couetoux2011rave}.
\subsection{Experimental results}
\begin{figure}[!tb]
  \centering
  \vspace*{5pt}
  \includegraphics[width=0.5\linewidth]{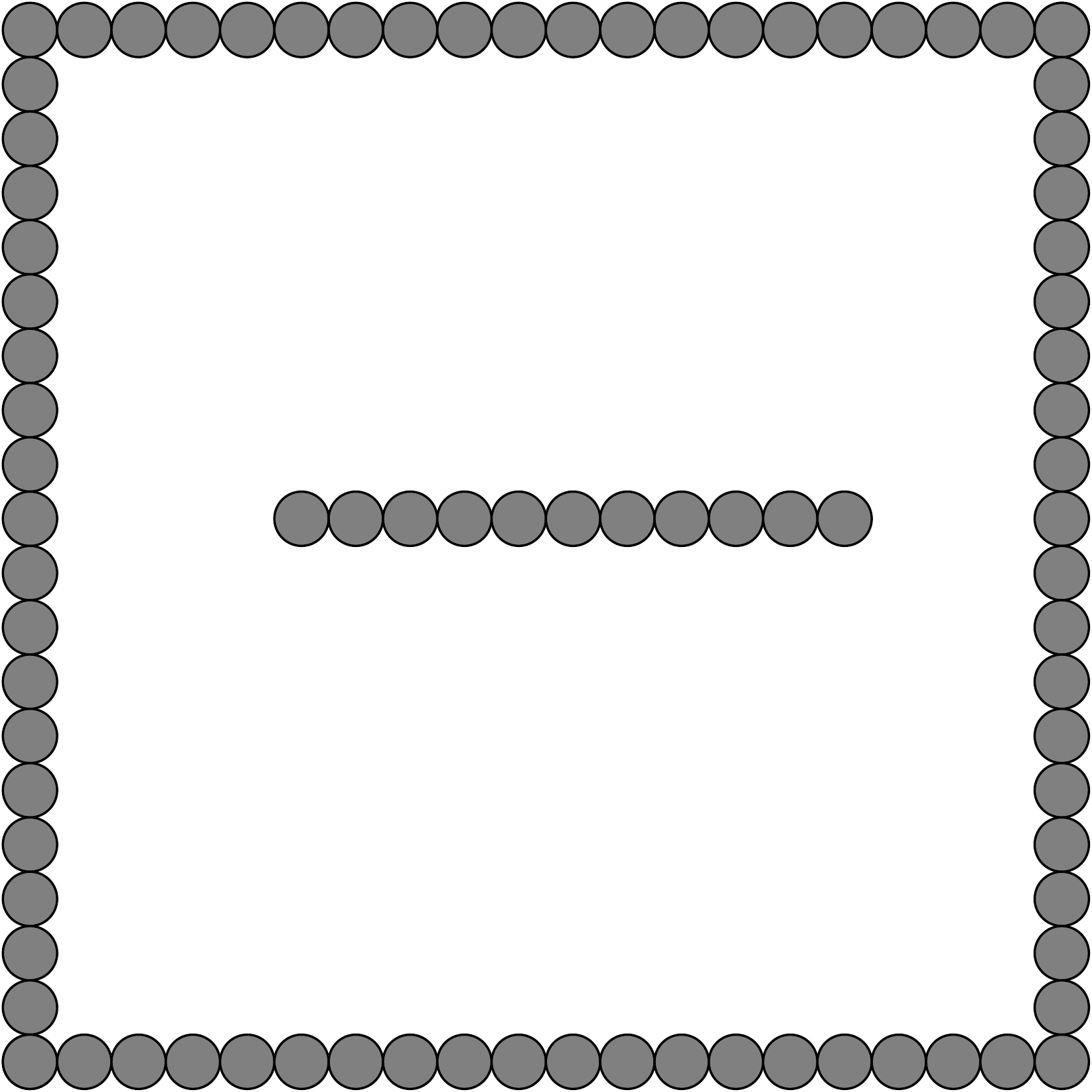}
  \caption{The environment used for the mobile robot task ($800\times 800$ units). The radii of the robot and the obstacles are the same (20 units).}
  \label{fig:mobile_map}
\end{figure}
\begin{figure}[!t]
  \centering
  \vspace*{5pt}
  \includegraphics[width=0.9\linewidth]{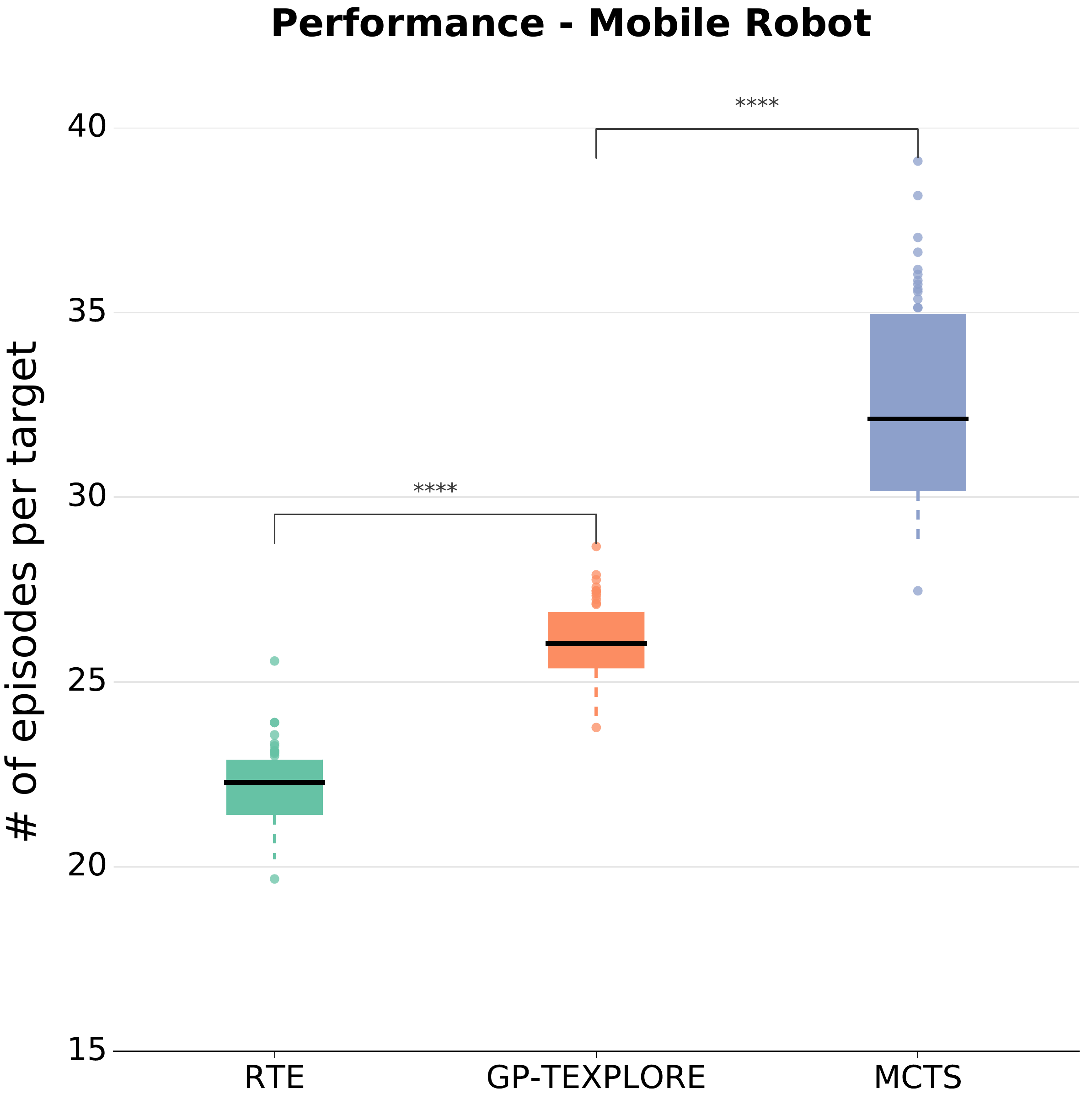}
  \caption{Comparison between RTE, GP-TEXPLORE and MCTS-based planning --- Differential drive robot simulation results. A damaged velocity-controlled differential drive robot (the right wheel's velocity command is halved) has to reach 30 random equidistant sequential targets. We replicated the scenario 50 times. RTE significantly outperforms (lower is better) both the re-planning baseline and GP-TEXPLORE. The number of stars indicates that the p-value of the Mann-Whitney U test is less than $0.05$, $0.01$, $0.001$ and $0.0001$ respectively.}
  \label{fig:sim_results_mobile}
\end{figure}
A damaged velocity-controlled differential drive robot (the right wheel's velocity command is halved) has to reach 30 random equidistant sequential targets in an arena with an obstacle in the middle (Fig.~\ref{fig:mobile_map}). The scenario is replicated 50 times for statistics.

We count the number of episodes (100 steps of simulation with the same velocity commands) required by the different algorithms to reach each target. The results show that RTE requires significantly fewer episodes ($22.28$ episodes, $25^{th}$ and $75^{th}$ percentiles $[21.4, 22.9]$) to reach each target than the re-planning baseline ($32.12$ episodes, $[30.17, 34.97]$) and GP-TEXPLORE ($26.03$ episodes, $[25.37, 26.9]$) (Fig.~\ref{fig:sim_results_mobile}).

\begin{figure}[!t]
  \centering
  \vspace*{5pt}
  \includegraphics[width=\linewidth]{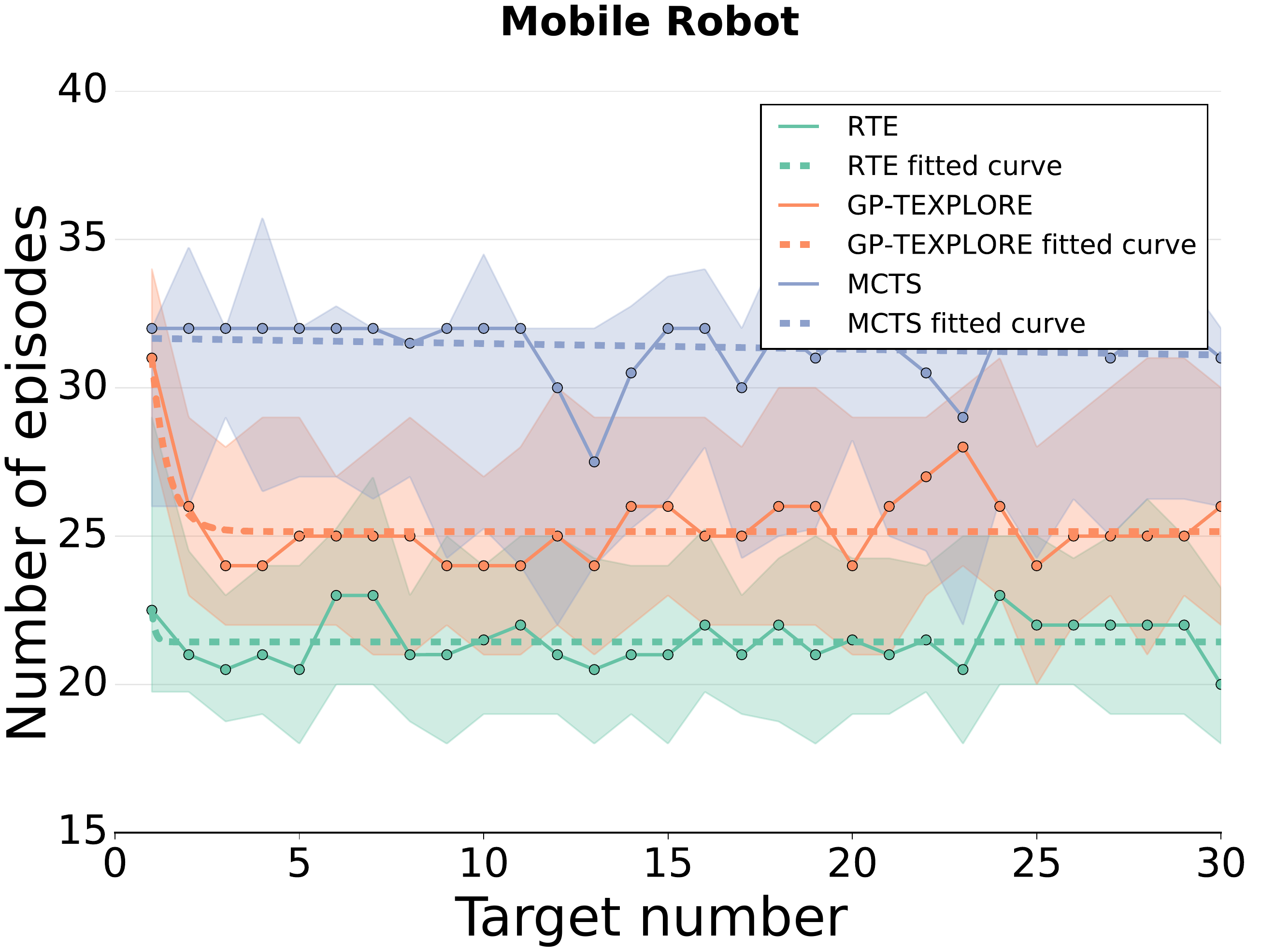}
  \caption{Median number of episodes to reach each target for a typical run of the algorithm for the mobile robot task. Over time, the robot using RTE or GP-TEXPLORE is able to reduce the number of required episodes to reach the next target (bottom lines), whereas MCTS alone uses a constant number of episodes (top line). Furthermore, RTE is able to correct its repertoire after the first target and outperforms GP-TEXPLORE, although the latter is capable of planning in the full action space. Most of the variance comes from the fact that the random targets are equidistant, but not of the same difficulty. The thick lines represent the medians over 50 runs and the shaded regions the $25^{th}$ and $75^{th}$ percentiles.}
  \label{fig:sim_results_line_mobile}
\end{figure}

Further analysis shows that the median number of episodes to reach each target decreases over time (until it reaches a steady value) when the robot uses RTE or GP-TEXPLORE, whereas it stays constant with MCTS alone (Fig.~\ref{fig:sim_results_line_mobile}). Furthermore, after the first target RTE is able to correct its repertoire and outperforms GP-TEXPLORE although the latter is capable of planning in the full action space.
\begin{table*}[!htb]
\footnotesize
\def\arraystretch{1.2}
\begin{center}
  \caption{Recovered locomotion capabilities - Wheeled Robot Task}
  \begin{tabulary}{\linewidth}{c|c|c|c|c|c|c}
   \hline
   \textbf{Intact} & \textbf{RTE} & \textbf{GP-TEXPLORE} & \textbf{MCTS} & \multicolumn{3}{c}{\textbf{Recovered capabilities}}\\
   \hline
   \multicolumn{4}{c|}{\textbf{Episodes per target}} & \textbf{RTE} & \textbf{GP-TEXPLORE} & \textbf{MCTS} \\
  \hline
   $14.08$ & $22.28$ & $26.03$ & $32.12$ & $63.20\%$ & $54.10\%$ & $43.85\%$ \\
   \hline
  \end{tabulary}
  \label{tab:mobile_recovered}
\end{center}
\end{table*}

We also performed the following evaluation test. We use the repertoire created by MAP-Elites with the intact robot and solve the same scenario (using MCTS as the planner --- no model learning, no variance). We replicate the scenario 50 times and take the median number of episodes required to reach a target. We then compute the percentage of the recovered capabilities using RTE, GP-TEXPLORE and MCTS-based planning. The results show that RTE recovers more locomotion capabilities than GP-TEXPLORE (Table~\ref{tab:mobile_recovered}); RTE is able to recover around $63\%$ of the original capabilities, whereas GP-TEXPLORE only recovers around $54\%$. Using only the repertoire generated with MAP-Elites and planning with MCTS is even worse, leading to only around $44\%$ of recovered capabilities. These results justify (1) that the repertoire itself is not enough for the robot to recover its abilities and (2) that using prior information (i.e., the repertoire) combined with learning (RTE) is beneficial compared to learning from scratch (GP-TEXPLORE).
\begin{figure}[!tb]
  \centering
  \vspace*{5pt}
  \includegraphics[width=0.8\linewidth]{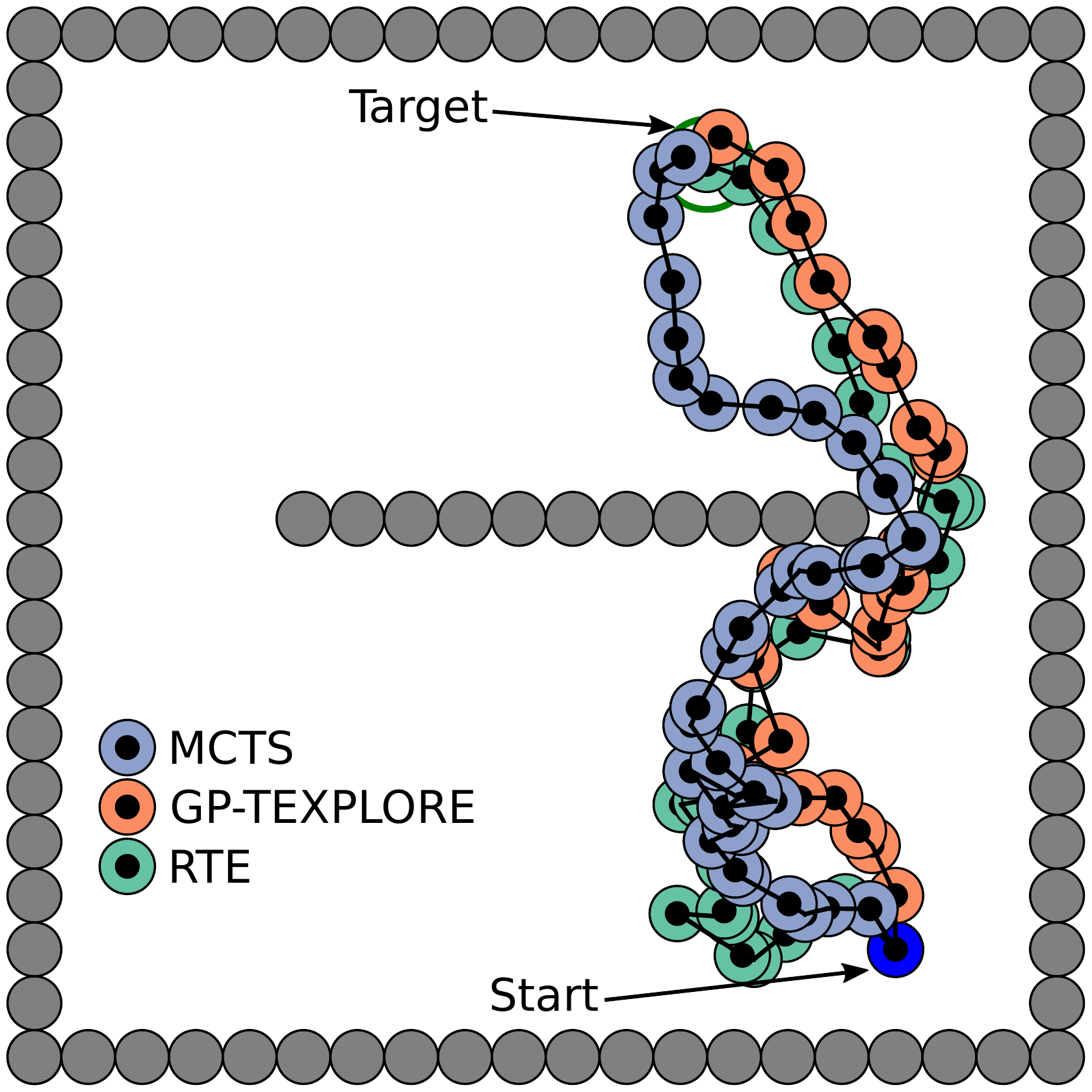}
  \caption{Sample trajectories of RTE, GP-TEXPLORE and MCTS in the mobile robot task. In this simple task, RTE and GP-TEXPLORE do not differ a lot (although RTE produces \emph{safer} and slightly faster paths --- Fig.\ref{fig:sim_results_mobile}) and produce higher performing paths than the MCTS baseline.}
  \label{fig:sim_traj_mobile}
\end{figure}

Finally, we observed that in this simple task, RTE and GP-TEXPLORE produce fairly similar paths, with the ones produced by RTE being slightly safer (i.e., not too close to the obstacles --- Fig.~\ref{fig:sim_traj_mobile}). In addition, both RTE and GP-TEXPLORE produce faster and safer paths than the MCTS baseline (Fig.~\ref{fig:sim_traj_mobile}). We also observed that the MCTS baseline often got stuck at the walls of the arena.
\section{Hexapod Robot Results}
\begin{figure*}[!t]
  \centering
  \vspace*{5pt}
  \includegraphics[width=0.75\linewidth]{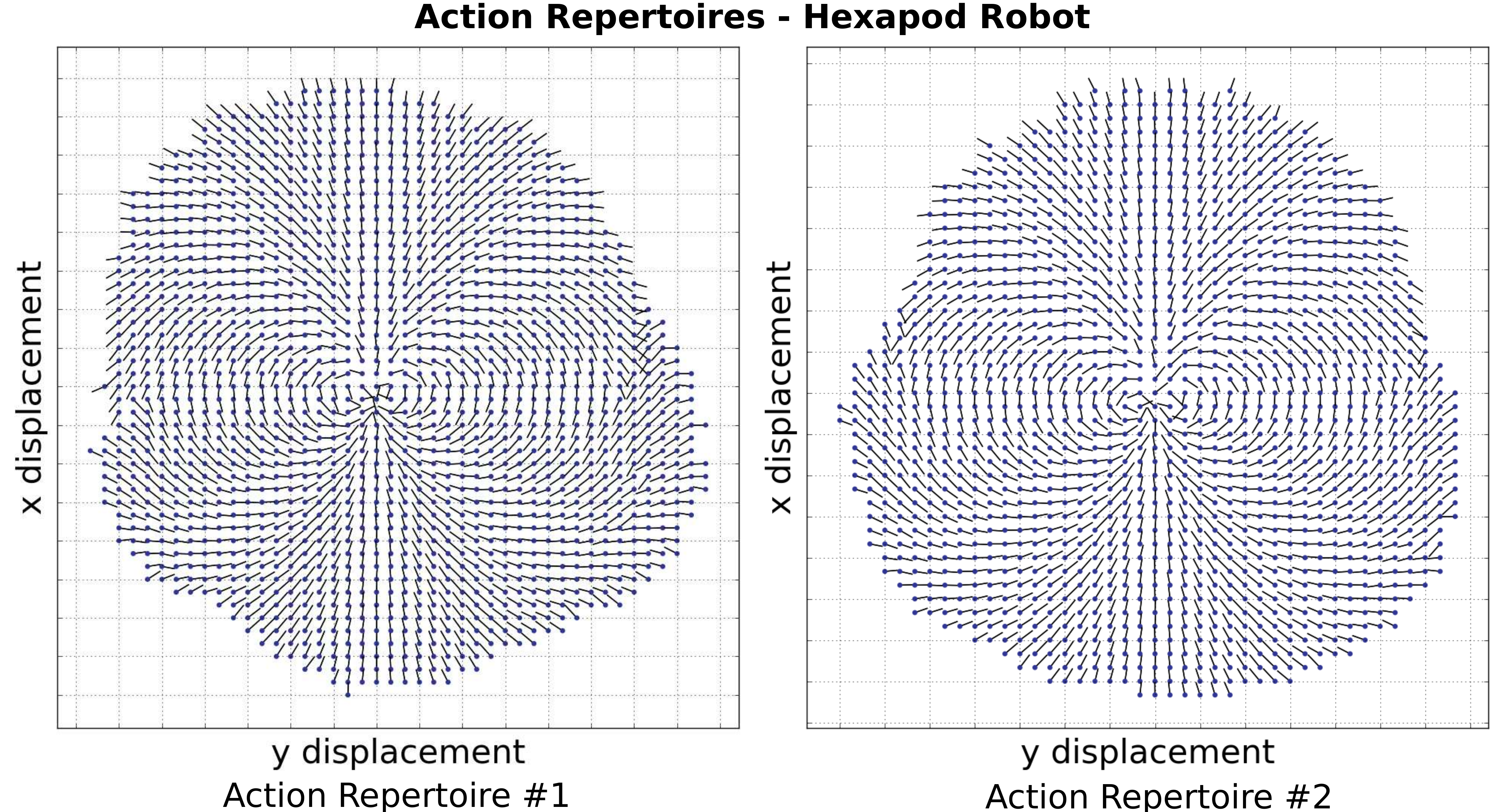}
  \caption{Repertoires for hexapod locomotion produced by the MAP-Elites algorithm. These repertoires map the 2-D action descriptor (of the 3-D task space) to the 36-D controller space. Each dot represents a different action (and its $x,y$ position), while the lines indicate the orientation of the robot at the end of each behavior. All the behaviors are relative to the zero position that is located in the middle of the figures and relative to the forward orientation (line pointing up).}
  \label{fig:map_repertoires}
\end{figure*}
Each leg of the hexapod robot that we used in our experiments has 3 degrees of freedom (DOF). This makes a total of 18-DOF for the whole robot. Nevertheless, since we are focusing on a path planning task, the state of the robot we are interested in consists of the $(x,y)$ position and the yaw angle $\theta$ of the center of mass (COM) of the robot, i.e., $\mathbf{s}_{hexa} = [x,y,\theta]$. The hexapod robot task and the simple mobile robot task share the same experimental setup and parameters, with the main differences between them being the following:
\begin{itemize}
  \item In order to produce periodic gaits for the hexapod, we do not control the robot in joint space, but use a low-level controller (Sec.~\ref{sec:controller}).
  \item The reachable space bounds for MAP-Elites are $[-2,2]$ meters and we set $l=0.03$ for the exponential kernel for the GP regression. In addition, to avoid depending on a specific repertoire, we ran MAP-Elites twice for $100000$ evaluations, leading to two distinct repertoires with about $1500$ different actions each (Fig.~\ref{fig:map_repertoires}). The hexapod is simulated using the {\it DART} simulator\footnote{\url{https://dartsim.github.io}}.
\end{itemize}
\subsection{Parametric Low-level Controller}
\label{sec:controller}
The low-level controller is the same as in~\cite{cully_robots_2015,cully_evolving_2015}. It is intentionally kept simple, so that this paper can focus on the learning algorithm. The angular position of each degree of freedom is governed by a periodic function $\Gamma$ parametrized by its amplitude $v$, its phase $\phi$, and its duty cycle $\tau$ (the duty cycle is the proportion of one period in which the joint is in its higher position). This function is a square signal of frequency 1Hz, amplitude $v$, and duty cycle $\tau$. A Gaussian filter is applied on the signal in order to remove sharp transitions, and it is then shifted according to the phase $\phi$.
The position of the third joint of each leg is the opposite of the position of the second one, so that the last segment is always vertical. This results in 36 real-valued parameters. Different values for these parameters can produce diverse gaits, from purely quadruped gaits to classic tripod gaits. At each \emph{episode} of the learning algorithm, the low-level controller is executed for 3 seconds with the specified parameters (for all algorithms).
\subsection{Simulation results}
\begin{figure*}[!t]
  \centering
  \vspace*{5pt}
  \includegraphics[width=0.9\linewidth]{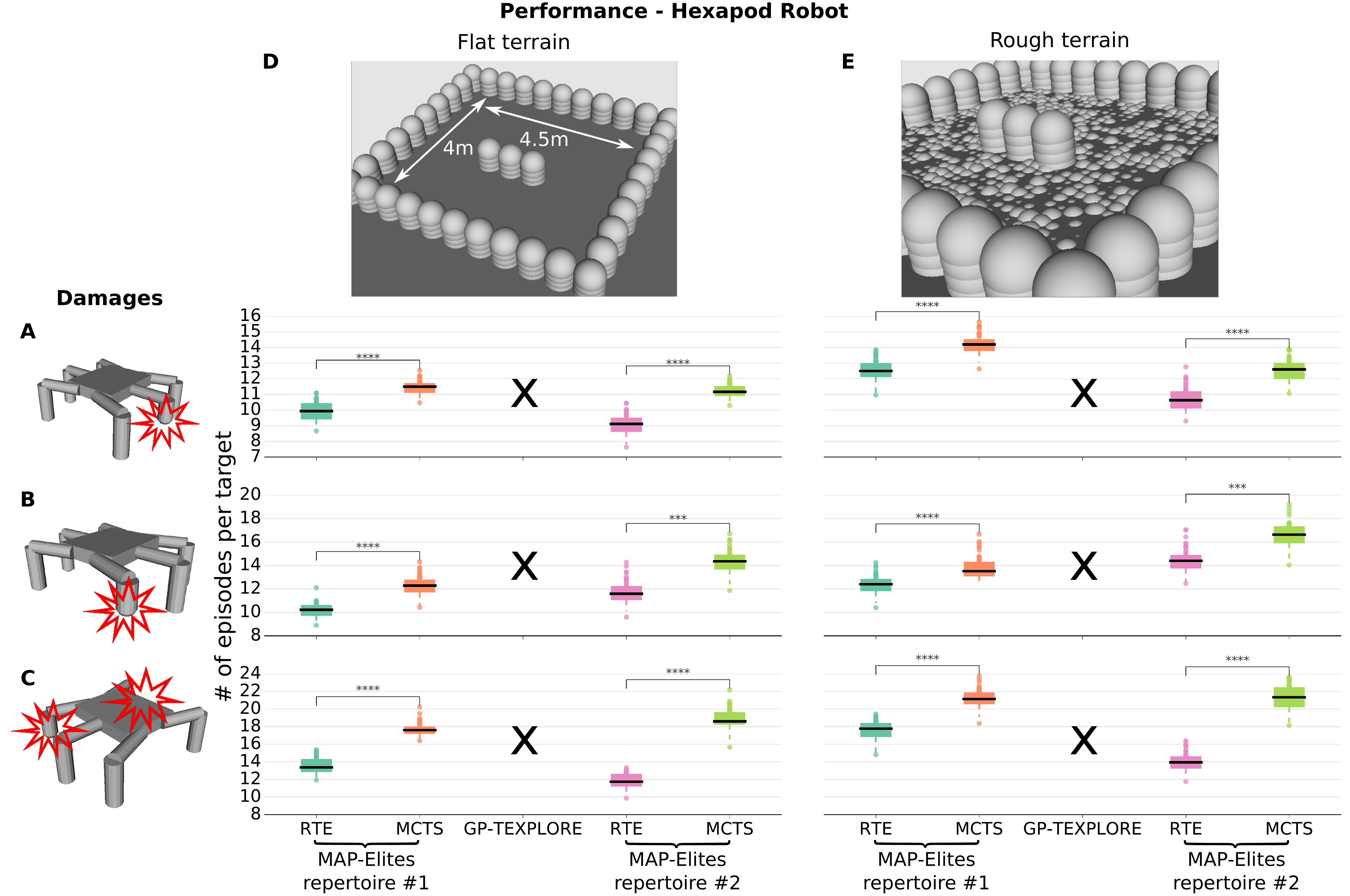}
  \caption{Comparison between RTE, GP-TEXPLORE and MCTS-based planning --- Hexapod robot simulation results. We investigate 3 different kinds of damage (\textbf{A} - middle leg shortening, \textbf{B} - back leg shortening, \textbf{C} - back leg shortening and middle leg removal), 2 different environments (\textbf{D} and \textbf{E}) and 2 different action repertoires. We replicated each scenario 50 times. The task is to reach 30 random equidistant sequential targets (distance of $3.5\,m$). RTE outperforms the re-planning baseline (lower is better). GP-TEXPLORE was not able to solve the task. The number of stars indicates that the p-value of the Mann-Whitney U test is less than $0.05$, $0.01$, $0.001$ and $0.0001$ respectively.}
  \label{fig:sim_results}
\end{figure*}

We count the number of episodes (3s actions) required to sequentially reach 30 equidistant (distance of $3.5\,m$) random targets. We investigate 3 different types of damage, 2 different environments (one with flat terrain and one with rough terrain), and 2 different action repertoires (Fig.~\ref{fig:sim_results}). Each scenario is replicated 50 times for statistics.

The results show that RTE requires significantly fewer episodes to reach each target than the re-planning baseline (Fig.~\ref{fig:sim_results}). Interestingly, RTE is able to reach the target points in the rough terrain scenario even though the action repertoire is learned on a flat terrain. This illustrates the capacity of RTE to compensate for unforeseen situations (i.e., damage and unmodeled terrain). Nevertheless, we observe slightly deteriorated performance and bigger differences between the MAP-Elites archives. This of course makes sense as the repertoires might have converged to different families of behaviors or one of them might be over-optimized for the flat terrain.

On the other hand, GP-TEXPLORE was not able to solve the task: with a budget of around 600 total episodes (due to computation time of GPs), it did not succeed in reaching a target (the robot would be reset to the next target position every 100 episodes). This is because learning a full dynamics model of a complex robot cannot be done with a few samples (i.e., less than 1000-2000~\cite{droniou2012learning}).

\begin{figure}[!t]
  \centering
  \vspace*{5pt}
  \includegraphics[width=\linewidth]{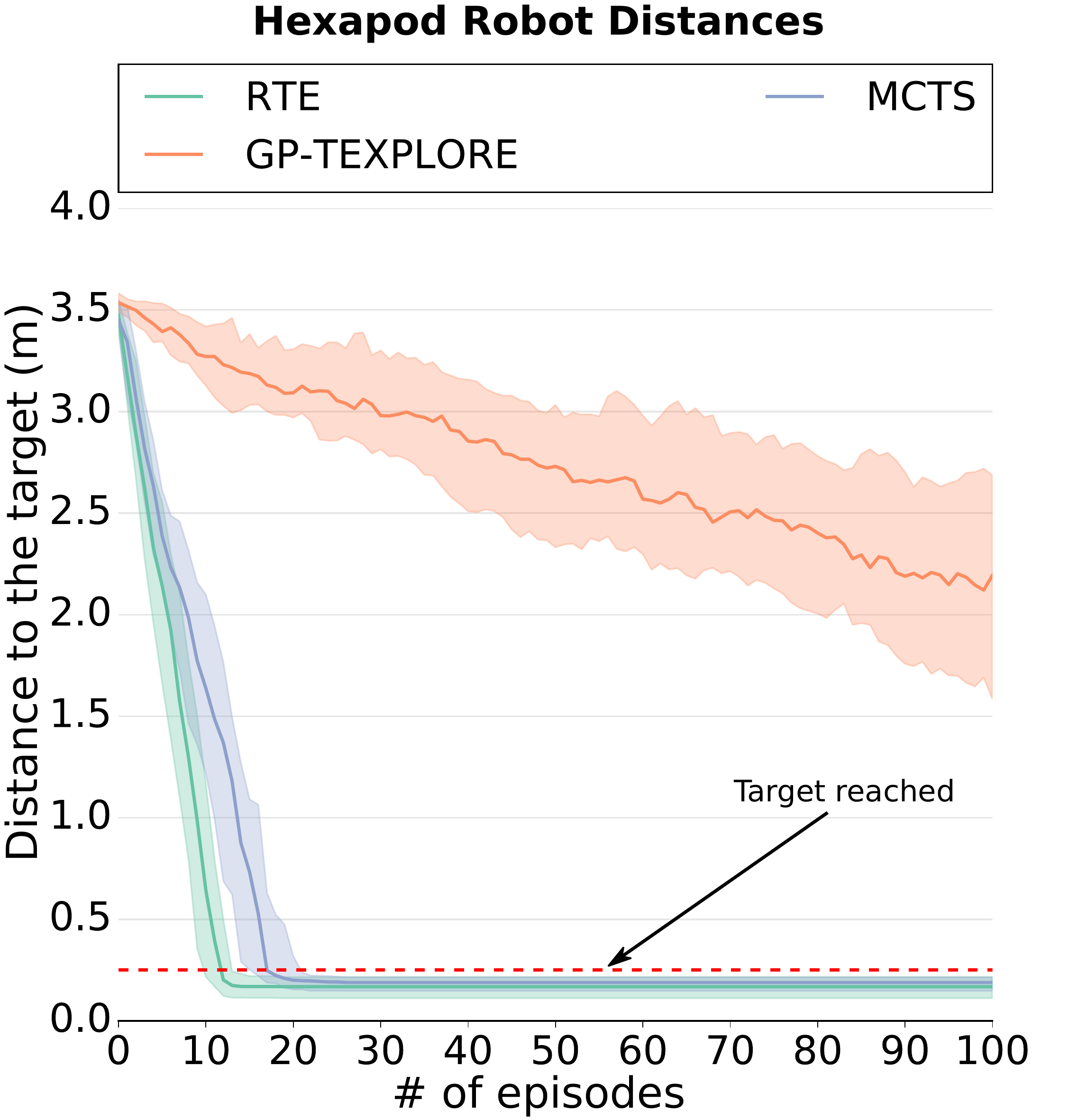}
  \caption{Comparison between RTE, GP-TEXPLORE and MCTS-based planning --- Hexapod robot simulation results. We measure the distance to the 5th target of RTE and GP-TEXPLORE as the number of episodes increases for the damage in Fig.~\ref{fig:sim_results}C, environment \#1 (Fig.~\ref{fig:sim_results}D) and the second repertoire. RTE clearly outperforms GP-TEXPLORE and the re-planning baseline; the robot with RTE reaches the target in about $10$ episodes, whereas with MCTS it needs more than $20$ episodes and with GP-TEXPLORE is not able to reach the target even after $100$ episodes. The lines represent medians over 50 runs and the shaded regions the $25^{th}$ and $75^{th}$ percentiles.}
  \label{fig:sim_distances}
\end{figure}

The results show that as the number of episodes increases, the robot that uses GP-TEXPLORE gets closer to the target, but cannot reach it when provided with a budget of 100 episodes (Fig.~\ref{fig:sim_distances}). On the contrary, the robot with RTE reaches the target in a small number of episodes (around 10 episodes in Fig.~\ref{fig:sim_distances}). Moreover, the robot that uses MCTS (the re-planning baseline) is still able to reach the target, but requires more episodes (around 20 episodes in Fig.~\ref{fig:sim_distances}). These results show that the pre-computed repertoire breaks the complexity of the problem and makes it tractable, but refining the repertoire is essential for damage recovery (or to handle the reality gap as illustrated below).

%
%
\begin{figure}[!t]
  \centering
  \vspace*{5pt}
  \includegraphics[width=\linewidth]{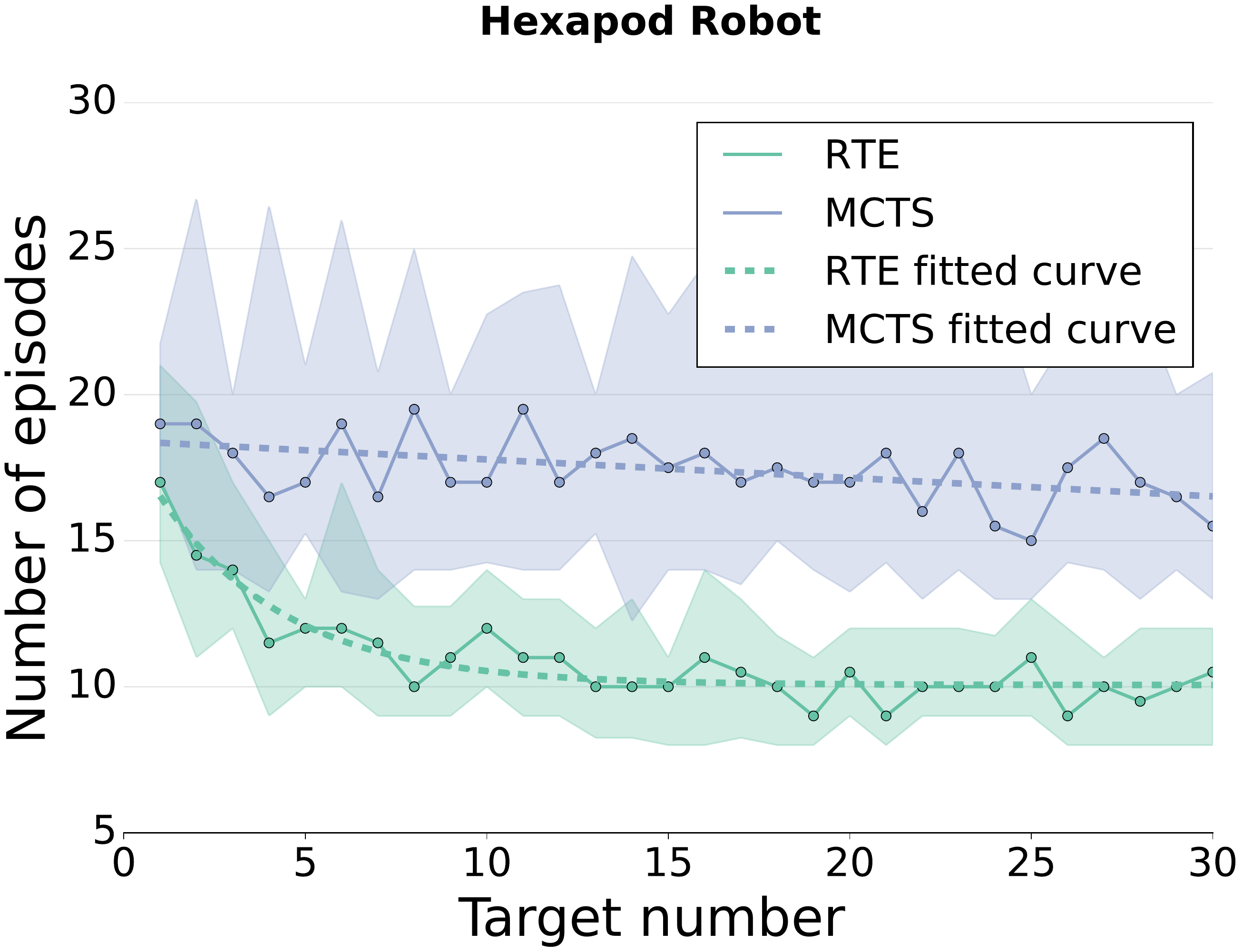}
  \caption{Median number of episodes to reach each target for a typical run of the algorithm in the hexapod task (in simulation --- for damage in Fig.~\ref{fig:sim_results}C, environment \#1 Fig.~\ref{fig:sim_results}D and the second action repertoire). Over time, the robot using RTE is able to reduce the number of required episodes to reach the next target (bottom line), whereas MCTS alone uses a constant number of episodes (top line). Most of the variance is due to the random targets being equidistant, but not of the same difficulty. The thick lines represent the medians over 50 runs and the shaded regions the $25^{th}$ and $75^{th}$ percentiles.}
  \label{fig:sim_results_line}
\end{figure}

Further analysis shows that the median number of episodes to reach each target decreases over time when the robot uses RTE, whereas it stays constant with MCTS alone (Fig.~\ref{fig:sim_results_line}). After the first few targets (2-4), RTE is able to make the robot reach each target in around $30s$ compared to MCTS alone that needs around $50-60\,s$.
\begin{table*}[!htb]
\scriptsize
\def\arraystretch{1.2}
\begin{center}
  \caption{Recovered locomotion capabilities - Hexapod Robot Task (Flat terrain scenarios)}
  \begin{tabulary}{\linewidth}{c|c||c|c|c|c|c}
   \hline
   \multicolumn{2}{c||}{\textbf{Flat terrain scenarios}} & \textbf{Intact} & \textbf{RTE} & \textbf{MCTS} & \multicolumn{2}{c}{\textbf{Recovered capabilities}}\\
   \hline
   \multicolumn{2}{c||}{}&\multicolumn{3}{c|}{\textbf{Episodes per target}} & \textbf{RTE} & \textbf{MCTS}\\
  \hline
   \multirow{3}{*}{Repertoire \#1} & Damage 1 (Fig.~\ref{fig:sim_results}A) & \multirow{3}{*}{$7.70$} & $9.93$ & $11.5$ & $77.52\%$ & $66.96\%$ \\
   & Damage 2 (Fig.~\ref{fig:sim_results}B) && $10.22$ & $12.28$ & $75.37\%$ & $62.69\%$ \\
   & Damage 3 (Fig.~\ref{fig:sim_results}C) && $13.37$ & $17.6$ & $57.61\%$ & $43.75\%$ \\
   \hline\hline
   \multirow{3}{*}{Repertoire \#2} & Damage 1 (Fig.~\ref{fig:sim_results}A) & \multirow{3}{*}{$7.12$} & $9.12$ & $11.17$ & $78.10\%$ & $63.76\%$ \\
   & Damage 2 (Fig.~\ref{fig:sim_results}B) && $11.58$ & $14.35$ & $61.44\%$ & $49.59\%$ \\
   & Damage 3 (Fig.~\ref{fig:sim_results}C) && $11.75$ & $18.6$ & $60.57\%$ & $38.26\%$ \\
   \hline
  \end{tabulary}
  \label{tab:hexa_recovered}
\end{center}
\end{table*}

\begin{table*}[!htb]
\scriptsize
\def\arraystretch{1.2}
\begin{center}
  \caption{Recovered locomotion capabilities - Hexapod Robot Task (Rough terrain scenarios)}
  \begin{tabulary}{\linewidth}{c|c||c|c|c|c|c}
   \hline
   \multicolumn{2}{c||}{\textbf{Rough terrain scenarios}} & \textbf{Intact} & \textbf{RTE} & \textbf{MCTS} & \multicolumn{2}{c}{\textbf{Recovered capabilities}}\\
   \hline
   \multicolumn{2}{c||}{}&\multicolumn{3}{c|}{\textbf{Episodes per target}} & \textbf{RTE} & \textbf{MCTS}\\
  \hline
   \multirow{3}{*}{Repertoire \#1} & Damage 1 (Fig.~\ref{fig:sim_results}A) & \multirow{3}{*}{$9.23$} & $12.5$ & $13.02$ & $73.84\%$ & $65\%$ \\
   & Damage 2 (Fig.~\ref{fig:sim_results}B) && $12.4$ & $13.52$ & $74.44\%$ & $68.29\%$ \\
   & Damage 3 (Fig.~\ref{fig:sim_results}C) && $17.78$ & $21.15$ & $51.90\%$ & $43.64\%$ \\
   \hline\hline
   \multirow{3}{*}{Repertoire \#2} & Damage 1 (Fig.~\ref{fig:sim_results}A) & \multirow{3}{*}{$8.55$} & $10.63$ & $12.60$ & $80.41\%$ & $67.86\%$ \\
   & Damage 2 (Fig.~\ref{fig:sim_results}B) && $14.4$ & $16.63$ & $59.38\%$ & $51.40\%$ \\
   & Damage 3 (Fig.~\ref{fig:sim_results}C) && $13.95$ & $21.33$ & $61.29\%$ & $40.10\%$ \\
   \hline
  \end{tabulary}
  \label{tab:hexa_recovered_rough}
\end{center}
\end{table*}

We also use the repertoire created by MAP-Elites with the intact robot to solve the same additional scenarios that were presented in the mobile robot case. We replicate the scenarios 50 times and take the median number of episodes required to reach a target. We then compute the percentage of the recovered capabilities using RTE and MCTS-based planning for all the damage conditions in the flat and the rough terrain environments.
The results show that RTE is able to almost always recover more than $60\%$ of the original capabilities (see Tables~\ref{tab:hexa_recovered} and~\ref{tab:hexa_recovered_rough}). These results are consistent with both the flat and the rough terrain evaluations. This demonstrates the robustness and the capacity of our approach to adapt to unforeseen situations. In addition, using the repertoire alone with MCTS planning is not enough for the robot to recover its capabilities as in half of the scenarios it fails to recover more than $60\%$ of the original capabilities and always recovers less than RTE.
\begin{figure}[!tb]
  \centering
  \vspace*{5pt}
  \includegraphics[width=0.8\linewidth]{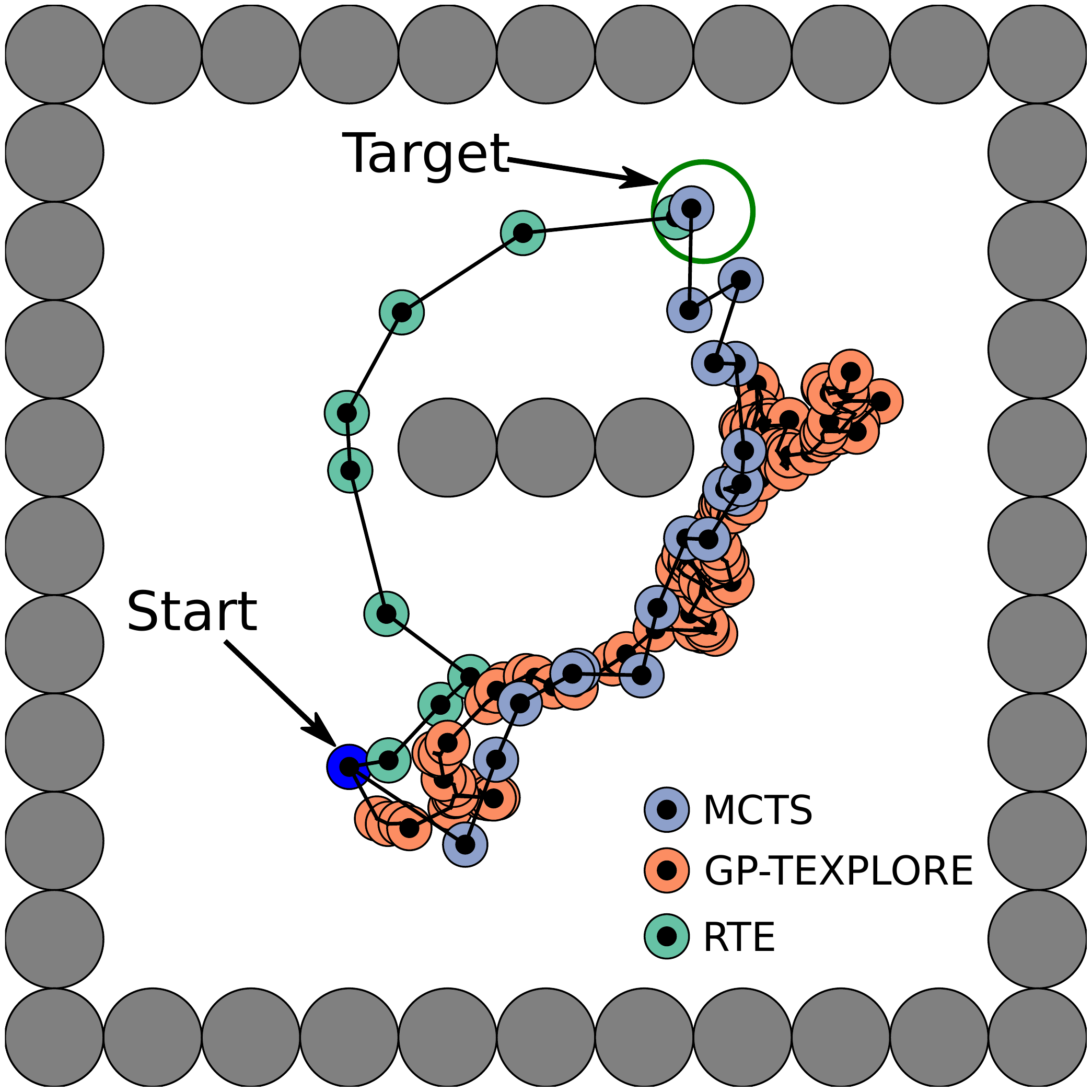}
  \caption{Sample trajectories of RTE, GP-TEXPLORE and MCTS in the simulated hexapod robot task. RTE produces faster and safer (i.e., not too close to the obstacles) paths than the MCTS baseline. GP-TEXPLORE cannot reach the target within a budget of 100 episodes, but does get closer to the target (as validated in~Fig.\ref{fig:sim_distances}).}
  \label{fig:sim_traj}
\end{figure}

Finally, we observed that RTE produces paths that are faster and safer than the MCTS baseline (Fig.~\ref{fig:sim_traj}). While GP-TEXPLORE cannot reach the target, it does get closer to the target point as the number of episodes increases (Fig.~\ref{fig:sim_traj} and Fig.~\ref{fig:sim_distances}). It is worth noting that GP-TEXPLORE takes actions that produce small displacements. This is probably due to the fact that the transition model cannot be accurately learned with a few data points, owing to the high dimensionality (36D) of the action space. As a consequence, the MCTS planner chooses actions that have already been selected. Since the search space is big and the first actions are selected almost randomly (there is no previous information), it is highly unlikely that taking these actions will actually lead to meaningful behaviors.
\subsection{Physical robot results}
\begin{figure}[!t]
  \centering
  \vspace*{5pt}
  \includegraphics[width=0.8\linewidth]{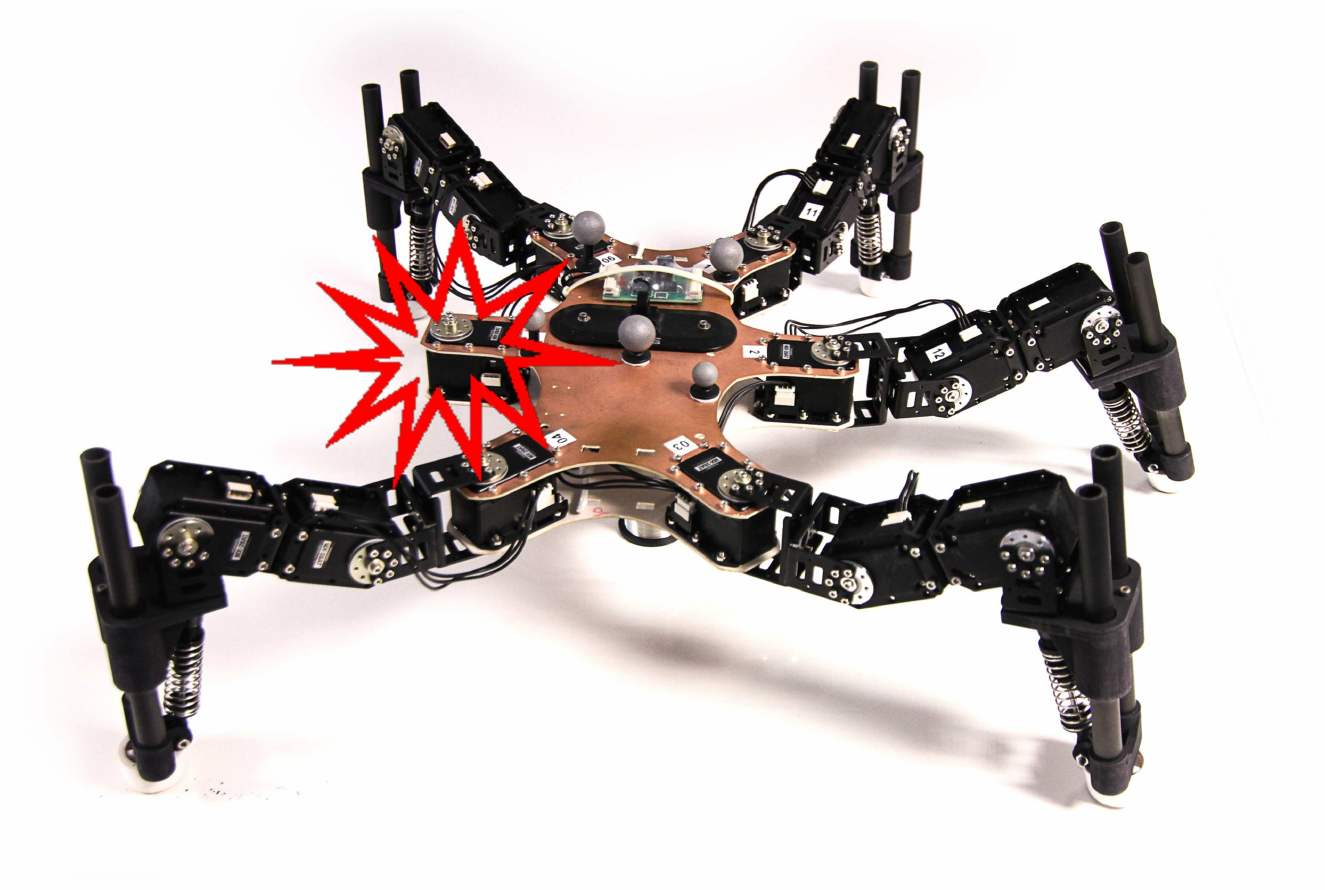}
  \caption{Physical damaged hexapod robot. The middle right leg is removed.}
  \label{fig:real_robot}
\end{figure}

\begin{figure*}[!t]
  \centering
  \vspace*{5pt}
  \includegraphics[width=0.8\linewidth]{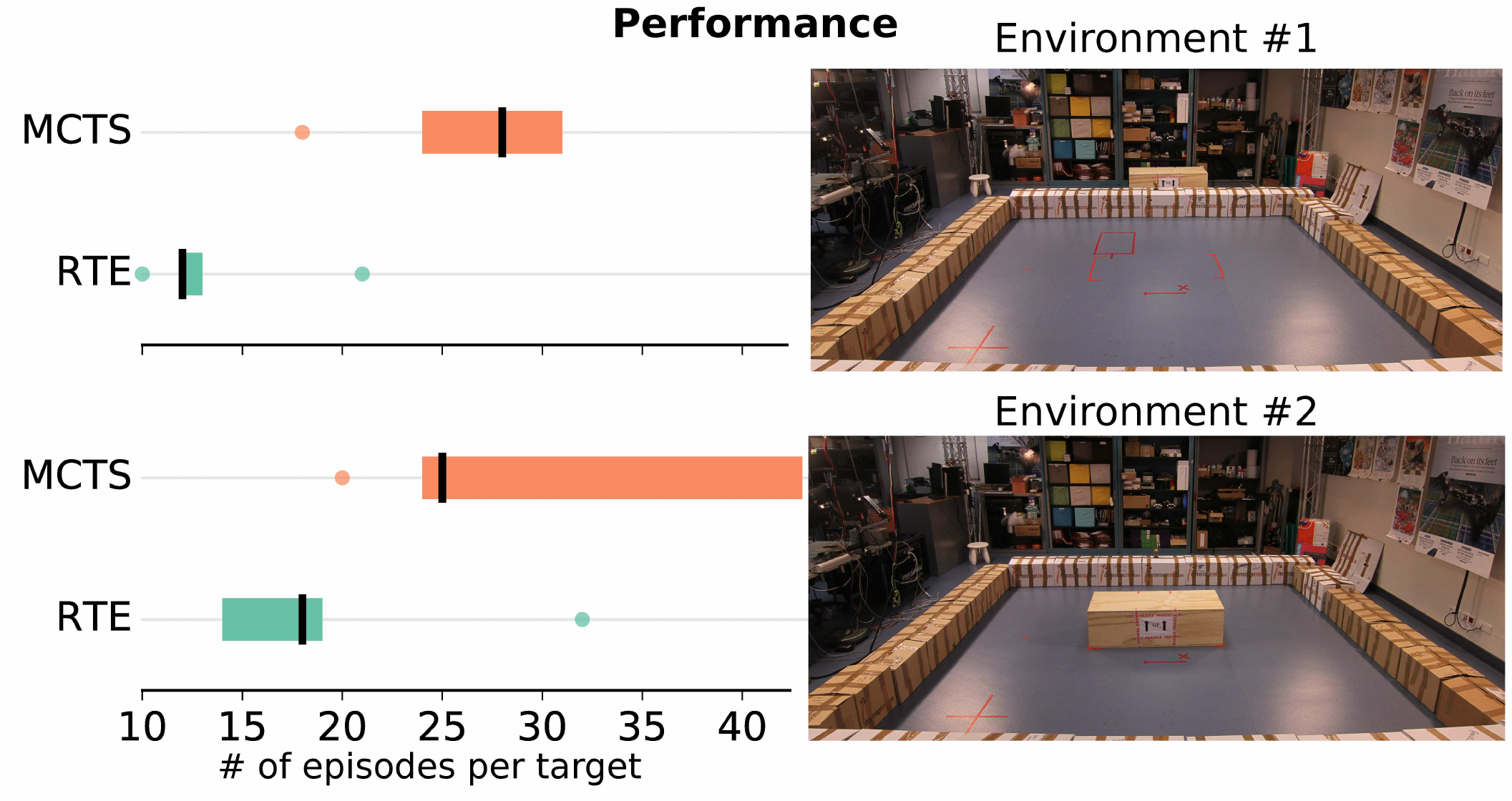}
  \caption{Comparison between RTE and MCTS-based planning --- Physical hexapod robot experiments. We investigate 1 damage, 2 different environments and 1 action repertoire. We replicated each scenario 5 times. The task is to reach 10 and 5 random equidistant ($2\sqrt{2}\,m$) sequential targets for the environment \#1 and \#2 respectively. RTE needs on average between $1.39$ and $2.33$ times fewer episodes to reach each target. The results are statistically significant (Mann-Whitney U test $p < 0.05$).}
  \label{fig:real_results}
\end{figure*}
We then evaluate RTE on the physical robot with a single damage (right middle leg removed --- Fig.~\ref{fig:real_robot}), in two environments (with and without a central obstacle) and the first action repertoire; the robot is required to reach 10 and 5 targets for each environment respectively, and the distance between the targets is $2\sqrt{2}\,m$. Each scenario is replicated 5 times. The environment (location of the obstacles) and the robot are tracked with an external motion capture system (Optitrack).

\begin{figure}[!tb]
  \centering
  \vspace*{5pt}
  \includegraphics[width=0.8\linewidth]{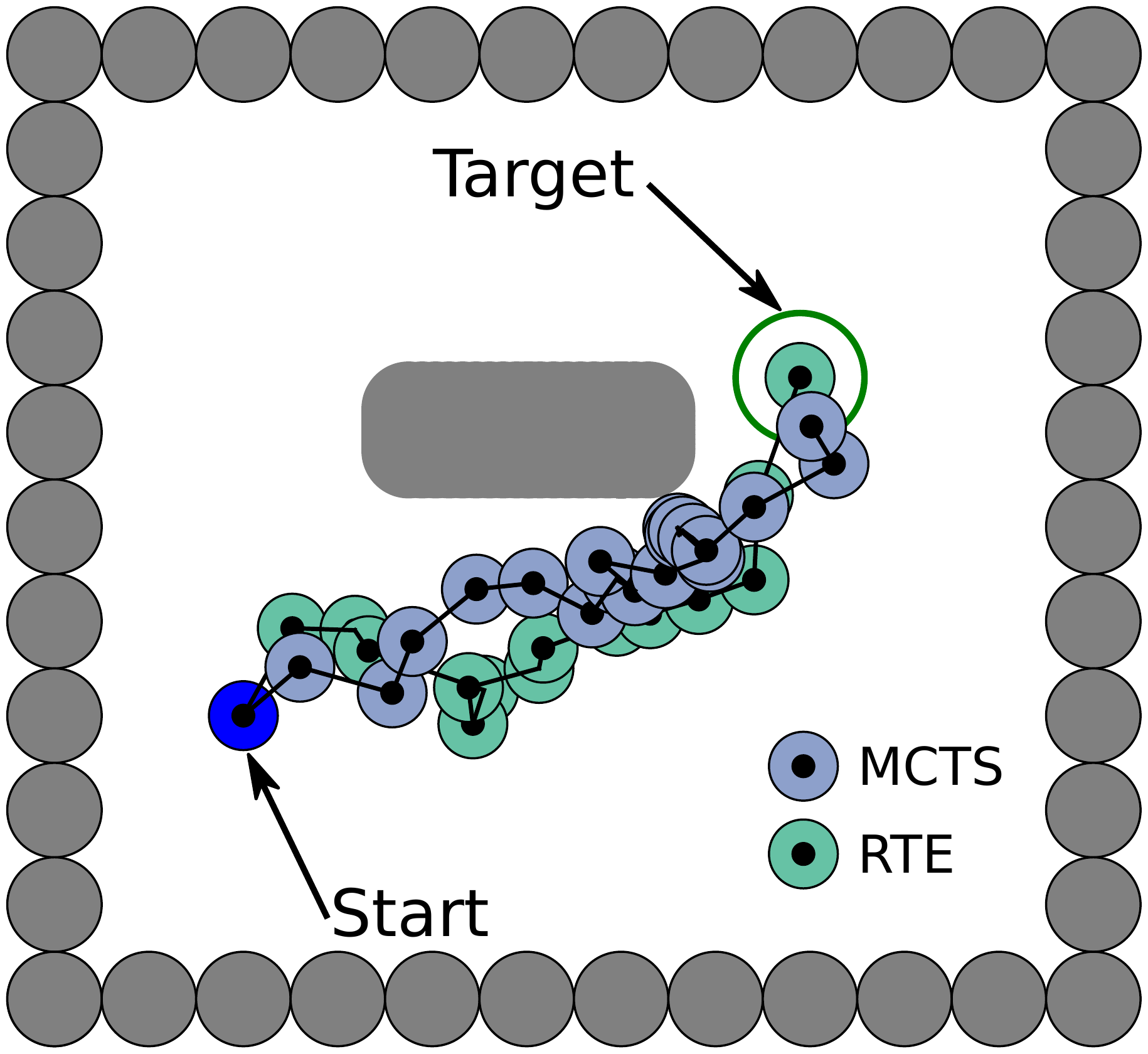}
  \caption{Sample trajectories of RTE and MCTS in the physical hexapod robot task. RTE comes up with faster and safer paths than the MCTS baseline to reach the goal point. The robot with the MCTS baseline tended to get stuck in the obstacle and struggle to get out of it and continue.}
  \label{fig:real_traj}
\end{figure}

The results show that RTE needs fewer episodes to reach each target (Environment 1: $13.0$ episodes, $25^{th}$ and $75^{th}$ percentiles $[12.0, 14.0]$, Environment 2: $18.0$ episodes, $[14.0, 19.0]$) than MCTS alone (Environment 1: $28.0$ episodes, $[24.0, 31.0]$, Environment 2: $25.0$ episodes, $[24.0, 43.0]$) (Fig.~\ref{fig:real_results}). These results are consistent with the simulations, but learning makes a bigger difference in the physical robot case. This is because the algorithm has to deal with the reality gap in addition to the damage in the physical robot case. Finally, as in the simulated experiments, RTE produces safer and faster paths than the MCTS baseline (Fig.~\ref{fig:real_traj}). The robot with the MCTS baseline tended to get stuck in the obstacle and struggle to get out of it and continue.
A demonstration of our approach on the real robot is available at \url{https://youtu.be/IqtyHFrb3BU}.

\section{Discussion and Conclusion}
With robots, like with many complex systems, ``we should not wonder \emph{if} some mishap may happen, but rather ask \emph{what} one will do about it when it occurs''~\cite{Corbato2007}. This advice is especially important if we want to be able to send advanced and expensive robots into dangerous places like a destroyed nuclear plant~\cite{guizzo2011fukushima}, even with tele-operated robots. In such situations, a damaged robot would greatly benefit from last-resort algorithms that would allow it to come back to its operators.

The RTE learning algorithm makes it possible for robots to overcome such failures without the need for resets and human intervention. We successfully tested it on a simple mobile robot and on a hexapod robot that were damaged in several ways. Unlike most previous work, our algorithm does not require the robot to be returned to the same position after each trial: the robot learns autonomously, while taking into account its environment (obstacles). To our knowledge, this learning algorithm is one of the first algorithms that allows a physical legged robot to learn to walk without any human intervention, especially when there are obstacles.

The main limitation of RTE is that it chooses the optimal action for the damaged robot \emph{among the actions that were found offline} with a different model. As a consequence, it is very likely that there exist better actions for the damaged robot in the full controller space, but RTE cannot use them. Nonetheless, this approximation seems to be sufficient in our experiments (i.e., the robot was able to complete its tasks) and it is one of the reasons why RTE scales significantly better than traditional RL approaches. In addition, it seems possible to periodically analyze the data collected (e.g., once a day), update the original simulation, and re-generate the repertoire.

It is important to highlight that RTE is not a policy search method, like PILCO~\cite{deisenroth_gaussian_2015} or Black-DROPS~\cite{chatzilygeroudis2017black}: RTE uses an approximate planner (MCTS) to derive a policy given the current model which, in turn, allows the robot to collect samples from the environment, refine the model, and thereby improve the policy, that is, the planner. Thus, the online phase of RTE can be seen as an on-policy, model-based RL procedure. In addition, the first phase of RTE (MAP-Elites), learns ``elementary behaviors'' (actions) in simulation, which are similar to parametrized policies or movement primitives \cite{ijspeert2002learning}. Nevertheless, the first phase of RTE does not only do that, but also creates a mapping from the high-dimensional controller space to the lower-dimensional task space, which proves to be beneficial when dealing with complex robots.
%

Ultimately, RTE should run continuously on the robot to compensate for potential wear or damage, that is, it should be a continuous learning, rather than a damage recovery, algorithm. However, the current version has a bottleneck: the computational complexity of the prediction of the GPs is cubic in the number of samples, which prevents the robot from using more than a few hundred episodes. A potential solution is reducing the query time of GPs by using a time-window and/or using sparse GPs~\cite{quinonero2005unifying} or local GPs~\cite{park2017patchwork}. Another solution is to replace the GPs with neural networks and take advantage of the recent advances in neural networks with uncertain predictions~\cite{gal2015dropout}.

In these first experiments, we assumed that the robot had perfect knowledge of its position and of the environment, which made it possible to cast our problem to an MDP. The next step is to relax this assumption and let the robot discover its environment with a SLAM algorithm~\cite{durrant2006simultaneous}. In this case, we could look at the problem from two different perspectives: (1) still treat the problem as an MDP and take into account the uncertainty of the map in the planning phase (MCTS), (2) treat the problem as a POMDP (Partially Observable MDP) and try to solve it with MCTS~\cite{silver2010monte}. The first perspective might not be enough to solve the problem (i.e., the robot would struggle to execute good plans), whereas the second one will increase the computation time of MCTS.


Furthermore, here we assumed that the outcome of each action is independent of the state it was taken in, which is the case for mobile robots when (1) the robot can be stopped to take a decision and (2) the terrain does not change dramatically. Nevertheless, RTE was able to cope with cases where this assumption did not really hold; in particular, the hexapod was able to walk on rough terrain, even though the action repertoire was optimized for flat terrain. In future work, we will look at this in greater depth, and try to relax these assumptions. For example, we could produce priors that are state-dependent and learn the full transition model and/or the reward function.

In this work, we chose to use MCTS for the planning phase of our approach, because it has been successfully used in the context of RL~\cite{silver2010monte,browne2012survey,hester2013texplore} and because it makes no assumptions about the dynamics/model of the system. This allows us to incorporate prior knowledge about the problem~\cite{silver2016mastering} and to use actions of any type, as we did in our work. Nevertheless, traditional sample-based planners, like RRT, could provide more accurate solutions and/or be faster in some cases. In future work, we will investigate and experiment with different probabilistic planners.


Lastly, while we performed our experiments with a legged and a mobile robot, the algorithm introduced here is general enough to be extended to many other robots and tasks. For instance, it could be employed on an arm mounted on a mobile platform that had incurred damage (e.g., a blocked joint). In this case, the algorithm will learn a mapping between the (x,y,z) position of the end-effector and the joint/wheel positions, similarly to how it learned a mapping between the (x,y) position of the hexapod robot and the parametric controller. After each trial, the robot might be in a different position relative to the target object (e.g., a door knob), but thanks to RTE, it should not have to go back to its starting position to try a different behavior.

\section*{Appendix}
For the mobile robot experiments, the threshold for reaching a target was $20$ units (the same as the radius of the robot). For the hexapod experiments, the threshold for reaching a target was $0.25\,m$ in simulation and $0.2\,m$ for the physical robot. The source code of the experiments can be found at {\footnotesize\url{https://github.com/resibots/chatzilygeroudis_2018_rte}}.%



\section*{Acknowledgments}
This work received funding from the European Research Council (ERC) under the European Union's Horizon 2020 research and innovation programme (grant agreement number 637972, project ``ResiBots''). The authors would like to thank Dorian Goepp, Antoine Cully, Olivier Buffet, Adam Gaier, and R\'emi Pautrat for their feedback.

%
%
%
%
%
\bibliography{mybib}

\begin{thebibliography}{10}
\expandafter\ifx\csname url\endcsname\relax
  \def\url#1{\texttt{#1}}\fi
\expandafter\ifx\csname urlprefix\endcsname\relax\def\urlprefix{URL }\fi
\expandafter\ifx\csname href\endcsname\relax
  \def\href#1#2{#2} \def\path#1{#1}\fi

\bibitem{atkeson_no_2015}
C.~Atkeson, et~al., No falls, no resets: Reliable humanoid behavior in the
  {DARPA} robotics challenge, in: Proc. of Humanoids, 2015, pp. 623--630.

\bibitem{carlson_how_2005}
J.~Carlson, R.~R. Murphy, How {UGVs} physically fail in the field, {IEEE}
  Trans. on Robotics 21~(3) (2005) 423--437.

\bibitem{dedonato2017team}
M.~DeDonato, F.~Polido, K.~Knoedler, B.~P. Babu, N.~Banerjee, C.~P. Bove,
  X.~Cui, R.~Du, P.~Franklin, J.~P. Graff, et~al., {Team WPI-CMU: Achieving
  Reliable Humanoid Behavior in the DARPA Robotics Challenge}, Journal of Field
  Robotics 34~(2) (2017) 381--399.

\bibitem{isermann2006fault}
R.~Isermann, Fault-diagnosis systems: an introduction from fault detection to
  fault tolerance, Springer Science \& Business Media, 2006.

\bibitem{verma_real-time_2004}
V.~Verma, G.~Gordon, R.~Simmons, S.~Thrun, Real-time fault diagnosis, {IEEE}
  Robotics \& Automation Magazine 11~(2) (2004) 56--66.

\bibitem{lengagne2013generation}
S.~Lengagne, J.~Vaillant, E.~Yoshida, A.~Kheddar, Generation of whole-body
  optimal dynamic multi-contact motions, International Journal of Robotics
  Research 32~(9-10) (2013) 1104--1119.

\bibitem{cully_robots_2015}
A.~Cully, J.~Clune, D.~Tarapore, J.-B. Mouret, Robots that can adapt like
  animals, Nature 521~(7553) (2015) 503--507.

\bibitem{koos_fast_2013}
S.~Koos, A.~Cully, J.-B. Mouret, Fast damage recovery in robotics with the
  {T}-resilience algorithm, International Journal of Robotics Research 32~(14)
  (2013) 1700--1723.

\bibitem{ren2015multiple}
G.~Ren, W.~Chen, S.~Dasgupta, C.~Kolodziejski, F.~W{\"o}rg{\"o}tter,
  P.~Manoonpong, Multiple chaotic central pattern generators with learning for
  legged locomotion and malfunction compensation, Information Sciences 294
  (2015) 666--682.

\bibitem{kober_reinforcement_2013}
J.~Kober, J.~A. Bagnell, J.~Peters, Reinforcement learning in robotics: {A}
  survey, International Journal of Robotics Research 32~(11) (2013) 1238--1274.

\bibitem{sutton1998reinforcement}
R.~S. Sutton, A.~G. Barto, Reinforcement learning: An introduction, MIT press,
  1998.

\bibitem{mnih_human_level_2015}
V.~Mnih, K.~Kavukcuoglu, D.~Silver, A.~A. Rusu, J.~Veness, M.~G. Bellemare,
  A.~Graves, M.~Riedmiller, A.~K. Fidjeland, G.~Ostrovski, S.~Petersen,
  C.~Beattie, A.~Sadik, I.~Antonoglou, H.~King, D.~Kumaran, D.~Wierstra,
  S.~Legg, D.~Hassabis, Human-level control through deep reinforcement
  learning, Nature 518~(7540) (2015) 529--533.

\bibitem{deisenroth2013survey}
M.~P. Deisenroth, G.~Neumann, J.~Peters, A survey on policy search for
  robotics., Foundations and Trends in Robotics 2~(1-2) (2013) 1--142.

\bibitem{ijspeert2002learning}
A.~J. Ijspeert, J.~Nakanishi, S.~Schaal, Learning attractor landscapes for
  learning motor primitives, in: Proc. of NIPS, 2002, pp. 1547--1554.

\bibitem{levine2013guided}
S.~Levine, V.~Koltun, Guided policy search, in: Proc. of ICML, no.~3 in {JMLR}
  Workshop and Conference Proceedings, 2013, pp. 1--9.

\bibitem{stulp_robot_2013}
F.~Stulp, O.~Sigaud, Robot skill learning: From reinforcement learning to
  evolution strategies, Paladyn, Journal of Behavioral Robotics 4~(1) (2013)
  49--61.

\bibitem{deisenroth_gaussian_2015}
M.~P. Deisenroth, D.~Fox, C.~E. Rasmussen, Gaussian processes for
  data-efficient learning in robotics and control, IEEE Trans. Pattern Anal.
  Mach. Intell. 37~(2) (2015) 408--423.

\bibitem{chatzilygeroudis2017black}
K.~Chatzilygeroudis, R.~Rama, R.~Kaushik, D.~Goepp, V.~Vassiliades, J.-B.
  Mouret, {Black-Box Data-efficient Policy Search for Robotics}, in: Proc. of
  IROS, 2017.

\bibitem{deisenroth_learning_2011}
M.~P. Deisenroth, C.~E. Rasmussen, D.~Fox, Learning to control a low-cost
  manipulator using data-efficient reinforcement learning, in: Robotics:
  Science \& Systems ({RSS}), 2011, pp. 57--64.

\bibitem{silver2016mastering}
D.~Silver, A.~Huang, C.~J. Maddison, A.~Guez, L.~Sifre, G.~Van Den~Driessche,
  J.~Schrittwieser, I.~Antonoglou, V.~Panneershelvam, M.~Lanctot, S.~Dieleman,
  D.~Grewe, J.~Nham, N.~Kalchbrenner, I.~Sutskever, T.~Lillicrap, M.~Leach,
  K.~Kavukcuoglu, T.~Graepel, D.~Hassabis, Mastering the game of {Go} with deep
  neural networks and tree search, Nature 529~(7587) (2016) 484--489.

\bibitem{chaslot_monte-carlo_2008}
G.~Chaslot, S.~Bakkes, I.~Szita, P.~Spronck, Monte-carlo tree search: A new
  framework for game {AI}, in: Proc. of {AIIDE}, 2008, pp. 216--217.

\bibitem{nguyen2011model}
D.~Nguyen-Tuong, J.~Peters, Model learning for robot control: a survey,
  Cognitive Processing 12~(4) (2011) 319--340.

\bibitem{hester2013texplore}
T.~Hester, P.~Stone, {TEXPLORE}: real-time sample-efficient reinforcement
  learning for robots, Machine Learning 90~(3) (2013) 385--429.

\bibitem{baranes2013active}
A.~Baranes, P.-Y. Oudeyer, Active learning of inverse models with intrinsically
  motivated goal exploration in robots, Robotics and Autonomous Systems 61~(1)
  (2013) 49--73.

\bibitem{nori2015icub}
F.~Nori, S.~Traversaro, J.~Eljaik, F.~Romano, A.~Del~Prete, D.~Pucci, {iCub
  whole-body control through force regulation on rigid non-coplanar contacts},
  Frontiers in Robotics and AI 2 (2015) 6.

\bibitem{peters2008reinforcement}
J.~Peters, S.~Schaal, Reinforcement learning of motor skills with policy
  gradients, Neural Networks 21~(4) (2008) 682--697.

\bibitem{2012ACLI2061}
J.-B. Mouret, S.~Doncieux, Encouraging behavioral diversity in evolutionary
  robotics: an empirical study, Evolutionary Computation 20~(1) (2012) 91--133.

\bibitem{calandra2015bayesian}
R.~Calandra, A.~Seyfarth, J.~Peters, M.~Deisenroth, Bayesian optimization for
  learning gaits under uncertainty, Annals of Mathematics and Artificial
  Intelligence 76 (2015) 5--23.

\bibitem{Lizotte2007}
D.~J. Lizotte, T.~Wang, M.~H. Bowling, D.~Schuurmans, Automatic gait
  optimization with gaussian process regression, in: Proc. of IJCAI, 2007, pp.
  944--949.

\bibitem{montgomery2016rfgps}
W.~Montgomery, A.~Ajay, C.~Finn, P.~Abbeel, S.~Levine, Reset-free guided policy
  search: Efficient deep reinforcement learning with stochastic initial states,
  arxiv:1610.01112.

\bibitem{tedrake2004stochastic}
R.~Tedrake, T.~W. Zhang, H.~S. Seung, Stochastic policy gradient reinforcement
  learning on a simple {3D} biped, in: Proc. of IROS, 2004, pp. 2849--2854.

\bibitem{peters2010relative}
J.~Peters, K.~M{\"u}lling, Y.~Altun, Relative entropy policy search, in: Proc.
  of AAAI, 2010, pp. 1607--1612.

\bibitem{schulman2015trust}
J.~Schulman, S.~Levine, P.~Moritz, M.~I. Jordan, P.~Abbeel, Trust region policy
  optimization, in: Proc. of ICML, 2015, pp. 1889--1897.

\bibitem{hester2012rtmba}
T.~Hester, M.~Quinlan, P.~Stone, {RTMBA: A real-time model-based reinforcement
  learning architecture for robot control}, in: Proc. of ICRA, IEEE, 2012, pp.
  85--90.

\bibitem{browne2012survey}
C.~B. Browne, et~al., A survey of monte carlo tree search methods, IEEE Trans.
  on Computational Intelligence and AI in Games 4~(1) (2012) 1--43.

\bibitem{droniou2012learning}
A.~Droniou, S.~Ivaldi, P.~Stalph, M.~Butz, O.~Sigaud, Learning velocity
  kinematics: Experimental comparison of on-line regression algorithms, in:
  Robotica, 2012, pp. 15--20.

\bibitem{blanke2003diagnosis}
M.~Blanke, J.~Schr{\"o}der, Diagnosis and fault-tolerant control, Vol. 115,
  Springer, 2003.

\bibitem{bongard2006resilient}
J.~C. Bongard, V.~Zykov, H.~Lipson, Resilient machines through continuous
  self-modeling, Science 314~(5802) (2006) 1118--1121.

\bibitem{mostafa2010alternative}
K.~Mostafa, C.~Tsai, I.~Her, Alternative gaits for multiped robots with leg
  failures to retain maneuverability, International Journal of Advanced Robotic
  Systems 7~(4) (2010) 31.

\bibitem{shahriari2016taking}
B.~Shahriari, K.~Swersky, Z.~Wang, R.~P. Adams, N.~de~Freitas, Taking the human
  out of the loop: A review of bayesian optimization, Proceedings of the IEEE
  104~(1) (2016) 148--175.

\bibitem{lavalle2006planning}
S.~M. LaValle, Planning algorithms, {Cambridge University Press}, 2006.

\bibitem{lavalle1998rapidly}
S.~M. LaValle, {Rapidly-exploring random trees: A new tool for path planning},
  Tech. Rep. TR 98-11, Computer Science Dept., Iowa State University (1998).

\bibitem{kavraki1996probabilistic}
L.~E. Kavraki, P.~Svestka, J.-C. Latombe, M.~H. Overmars, Probabilistic
  roadmaps for path planning in high-dimensional configuration spaces, IEEE
  Trans. on Robotics and Automation 12~(4) (1996) 566--580.

\bibitem{mouret_illuminating_2015}
J.-B. Mouret, J.~Clune, Illuminating search spaces by mapping elites,
  arxiv:1504.04909.

\bibitem{cully_evolving_2015}
A.~Cully, J.-B. Mouret, Evolving a behavioral repertoire for a walking robot,
  Evolutionary Computation.

\bibitem{duarte2017evolution}
M.~Duarte, J.~Gomes, S.~M. Oliveira, A.~L. Christensen, {Evolution of
  repertoire-based control for robots with complex locomotor systems}, IEEE
  Trans. on Evolutionary Computation.

\bibitem{cully2017quality}
A.~Cully, Y.~Demiris, {Quality and Diversity Optimization: A Unifying Modular
  Framework}, IEEE Trans. on Evolutionary Computation.

\bibitem{duarte2016evorbc}
M.~Duarte, J.~Gomes, S.~M. Oliveira, A.~L. Christensen, {EvoRBC:} evolutionary
  repertoire-based control for robots with arbitrary locomotion complexity, in:
  Proc. of GECCO, 2016, pp. 93--100.

\bibitem{pugh2016quality}
J.~K. Pugh, L.~B. Soros, K.~O. Stanley, Quality diversity: A new frontier for
  evolutionary computation, Frontiers in Robotics and AI 3 (2016) 40, doi:
  10.3389/frobt.2016.00040.

\bibitem{gaier2017feature}
A.~Gaier, A.~Asteroth, J.-B. Mouret, Feature space modeling through surrogate
  illumination, in: Proc. of GECCO, 2017.

\bibitem{nguyen2015deep}
A.~Nguyen, J.~Yosinski, J.~Clune, Deep neural networks are easily fooled: High
  confidence predictions for unrecognizable images, in: Proc. of CVPR, 2015,
  pp. 427--436.

\bibitem{nguyen2016understanding}
A.~Nguyen, J.~Yosinski, J.~Clune, {Understanding Innovation Engines: Automated
  Creativity and Improved Stochastic Optimization via Deep Learning},
  Evolutionary Computation 24 (2016) 545--572.

\bibitem{lehman2016iccc}
J.~Lehman, S.~Risi, J.~Clune, Creative generation of {3D} objects with deep
  learning and innovation engines, in: Proc. of the 7th Intern. Conf. on
  Comput. Creativity, 2016, pp. 180--187.

\bibitem{vassiliades2017using}
V.~Vassiliades, K.~Chatzilygeroudis, J.-B. Mouret, {Using Centroidal Voronoi
  Tessellations to Scale Up the Multi-dimensional Archive of Phenotypic Elites
  Algorithm}, IEEE Trans. on Evolutionary Computation.

\bibitem{rasmussen2006gaussian}
C.~E. Rasmussen, C.~K.~I. Williams, Gaussian processes for machine learning,
  MIT Press, 2006.

\bibitem{silver2010monte}
D.~Silver, J.~Veness, Monte-carlo planning in large {POMDPs}, in: Proc. of
  NIPS, 2010, pp. 2164--2172.

\bibitem{couetoux2011continuous}
A.~Cou{\"e}toux, J.-B. Hoock, N.~Sokolovska, O.~Teytaud, N.~Bonnard, Continuous
  upper confidence trees, in: Proc. of {LION}, 2011, pp. 433--445.

\bibitem{mouret2010sferes}
J.-B. Mouret, S.~Doncieux, Sferes$_{v2}$: Evolvin'in the multi-core world, in:
  Proc. of IEEE CEC, 2010.

\bibitem{cully2016limbo}
A.~Cully, K.~Chatzilygeroudis, F.~Allocati, J.-B. Mouret, {Limbo: A Fast and
  Flexible Library for Bayesian Optimization}, arxiv:1611.07343.

\bibitem{rolet2009optimal}
P.~Rolet, M.~Sebag, O.~Teytaud, Boosting active learning to optimality: A
  tractable monte-carlo, billiard-based algorithm, in: Proc. of ECML, 2009, pp.
  302--317.

\bibitem{cazenave2007parallelization}
T.~Cazenave, N.~Jouandeau, {On the parallelization of UCT}, in: Proc. of the
  Computer Games Workshop, 2007, pp. 93--101.

\bibitem{couetoux2011rave}
A.~Couetoux, M.~Milone, M.~Brendel, H.~Doghmen, M.~Sebag, O.~Teytaud,
  Continuous rapid action value estimates, in: Proc. of ACML, 2011, p. 19–31.

\bibitem{Corbato2007}
F.~Corbato, {On Building Systems That Will Fail}, ACM Turing award lectures
  34~(9) (2007) 72--81.

\bibitem{guizzo2011fukushima}
E.~Guizzo, Fukushima robot operator writes tell-all blog, in: IEEE Spectrum,
  2011, {URL:
  \url{http://spectrum.ieee.org/automaton/robotics/industrial-robots/fukushima-robot-operator-diaries}}.

\bibitem{quinonero2005unifying}
J.~Qui{\~n}onero-Candela, C.~E. Rasmussen, {A unifying view of sparse
  approximate Gaussian process regression}, Journal of Machine Learning
  Research 6 (2005) 1939--1959.

\bibitem{park2017patchwork}
C.~Park, D.~Apley, Patchwork kriging for large-scale {Gaussian} process
  regression, arXiv preprint arXiv:1701.06655.

\bibitem{gal2015dropout}
Y.~Gal, Z.~Ghahramani, Dropout as a {Bayesian} approximation: Representing
  model uncertainty in deep learning, in: Proc. of ICML, 2016, pp. 1050--1059.

\bibitem{durrant2006simultaneous}
H.~Durrant-Whyte, T.~Bailey, Simultaneous localization and mapping: part {I},
  IEEE Robotics \& Automation Magazine 13~(2) (2006) 99--110.

\end{thebibliography}

\end{document}